\documentclass{article}

\usepackage{arxiv}

\usepackage[utf8]{inputenc} 
\usepackage[T1]{fontenc}    
\usepackage{hyperref}       
\usepackage{url}            
\usepackage{booktabs}       
\usepackage{amsfonts}       
\usepackage{nicefrac}       
\usepackage{microtype}      
\usepackage{cleveref}       
\usepackage{lipsum}         
\usepackage{graphicx}
\usepackage{natbib}
\usepackage{doi}

\usepackage{lineno}
\modulolinenumbers[5]
\usepackage{epsfig}
\usepackage{amssymb}
\usepackage{xcolor}
\usepackage{wasysym}
\usepackage[ruled,vlined]{algorithm2e}
\usepackage{subfigure}
\usepackage{amsfonts}
\usepackage{booktabs}
\usepackage{siunitx}
\usepackage{array}
\usepackage{hyperref}

\title{Unsupervised Segmentation for Terracotta Warrior Point Cloud (SRG-Net)}

\author{ \href{https://orcid.org/0000-0002-4611-1983}{\includegraphics[scale=0.06]{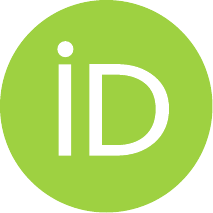}\hspace{1mm}Yao Hu} \\
	School of Information Science and Technology\\
	Northwest University\\
	Xi'an 710127, China \\
	\texttt{yhu6589@outlook.com} \\
	\And
	\href{https://orcid.org/0000-0002-4234-2119}{\includegraphics[scale=0.06]{orcid.pdf}\hspace{1mm}Guohua Geng}\thanks{Corresponding author.} \\
	School of Information Science and Technology\\
	Northwest University\\
	Xi'an 710127, China \\
	\texttt{ghgeng@nwu.edu.cn} \\
	\And
	\href{https://orcid.org/
0000-0001-6218-5715}{\includegraphics[scale=0.06]{orcid.pdf}\hspace{1mm}Kang Li} \\
	School of Information Science and Technology\\
	Northwest University\\
	Xi'an 710127, China \\
	\texttt{likang@nwu.edu.cn}
	\And
	Wei Zhou \\
	School of Information Science and Technology\\
	Northwest University\\
	Xi'an 710127, China \\
	\texttt{zhouwei@nwu.edu.cn}
}

\renewcommand{\headeright}{}
\renewcommand{\undertitle}{}

\hypersetup{
pdftitle={Unsupervised Segmentation for Terracotta Warrior Point Cloud (SRG-Net)},
pdfauthor={Yao Hu},
pdfkeywords={Point Cloud, Unsupervised Segmentation, Terracotta Warrior, Convolutional Neural Network, Seed Region Growing},
}

\begin{document}
\maketitle

\begin{abstract}
The repairing work of terracotta warriors in Emperor Qinshihuang Mausoleum Site Museum is handcrafted by experts, and the increasing amounts of unearthed pieces of terracotta warriors make the archaeologists too challenging to conduct the restoration of terracotta warriors efficiently. We hope to segment the 3D point cloud data of the terracotta warriors automatically and store the fragment data in the database to assist the archaeologists in matching the actual fragments with the ones in the database, which could result in higher repairing efficiency of terracotta warriors. Moreover, the existing 3D neural network research is mainly focusing on supervised classification, clustering, unsupervised representation, and reconstruction. There are few pieces of researches concentrating on unsupervised point cloud part segmentation. In this paper, we present SRG-Net for 3D point clouds of terracotta warriors to address these problems. Firstly, we adopt a customized seed-region-growing algorithm to segment the point cloud coarsely. Then we present a supervised segmentation and unsupervised reconstruction networks to learn the characteristics of 3D point clouds. Finally, we combine the SRG algorithm with our improved CNN(convolution neural network) using a refinement method. This pipeline is called SRG-Net, which aims at conducting segmentation tasks on the terracotta warriors. Our proposed SRG-Net is evaluated on the terracotta warrior data and ShapeNet dataset by measuring the accuracy and the latency. The experimental results show that our SRG-Net outperforms the state-of-the-art methods. Our code is available at https://github.com/hyoau/SRG-Net.
\end{abstract}


\section{Introduction} \label{introduction}

Nowadays, the repairing work of terracotta warriors is mainly accomplished by the handwork of archaeologists in Emperor Qinshihuang's Mausoleum Site Museum.
While more and more fragments of terracotta warriors are excavated from the site, the restoration does not have a high repairing efficiency. Moreover, the lack of archaeological technicians repairing the terracotta warriors makes it more uneasy to catch up with excavation speed.
As a result, more and more fragments excavated from the archaeological site remain to be repaired.
Terracotta Warrior is a 3D natural structure and can be represented with a point cloud.
Fortunately, with the development of 3D sensors (Apple Li-DAR, Kinect, and RealSense) and the improvement of hardware (GPU and CPU), it is possible to improve the repairing speed digitally. 
We can segment the 3D object of the terracotta warriors and store the fragments in the database. Our work aims at simplifying the work of researchers in the museum, making it easy to match fragments excavated from the site with the pre-segmented fragments in the database. 
To improve {the} efficiency of the repairing work, we propose an unsupervised 3D point cloud segmentation method for the terracotta warrior data based on convolution neural network. 
With the segmentation method proposed in this paper, the complete terracotta warriors can be segmented into fragments corresponding to various parts automatically, e.g., head, hand, and foot, etc.
Compared with the traditional segmentation work by hand, the method proposed in this paper can significantly improve efficiency and accuracy.

\begin{figure*}[h]
    \centering
	\includegraphics[width=0.8\textwidth]{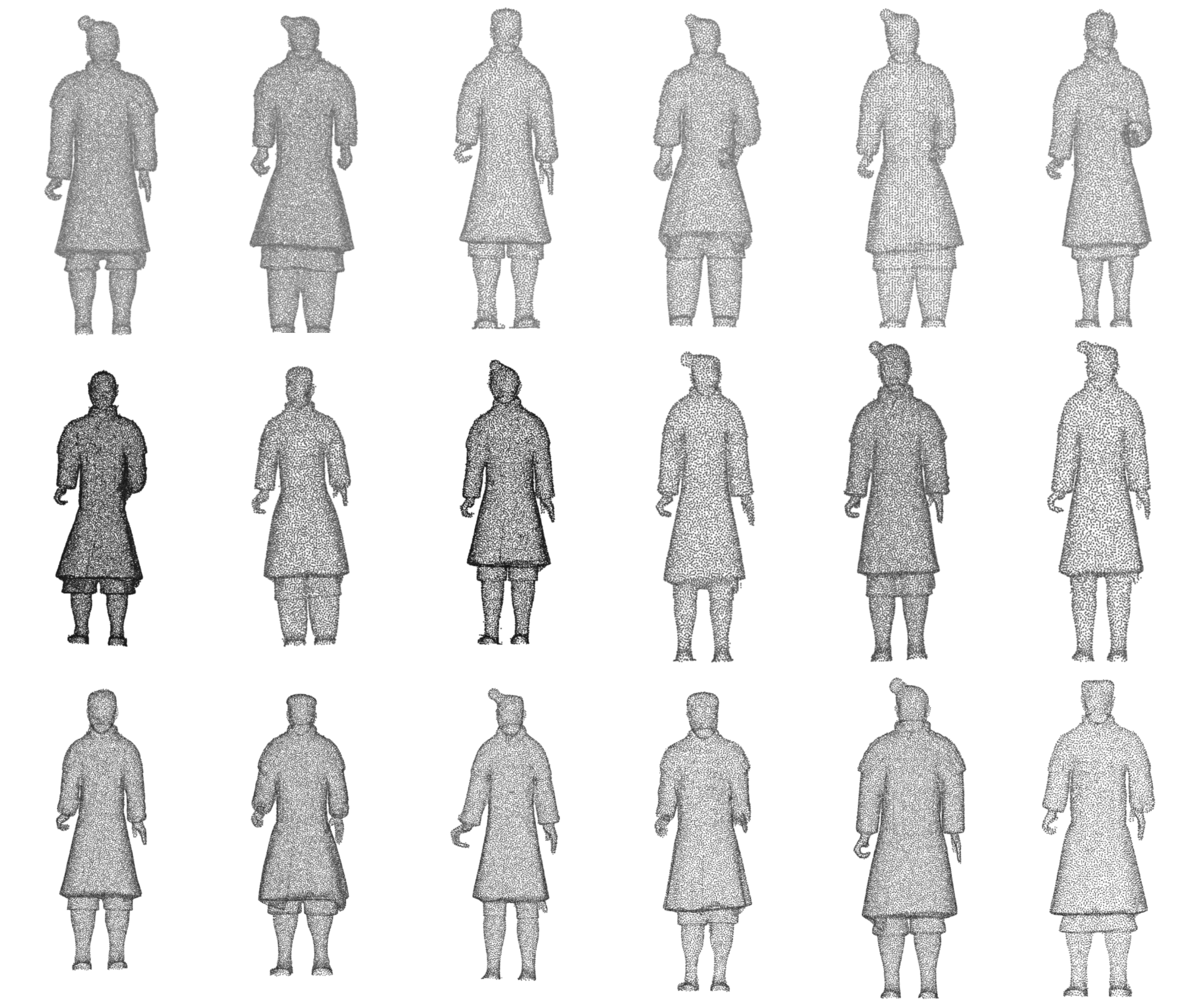}
	\caption{Simplified Terracotta Warrior Point Clouds}
	\label{figure demo}
\end{figure*}

Recently, there have been many pieces of research focusing on how to voxelize the point cloud to make them {evenly distributed} in regular 3D space, and then implement 3D-CNN on them.
However, voxelization owns the high space and time complexity, and there may be quantization errors {in the process of} voxelization, which would result in low accuracy. 
Compared with other data formats, point cloud is a data structure suitable for 3D scene calculation of terracotta warriors. We choose to segment {the terracotta warriors} in the form of point clouds (see samples in Fig.~\ref{figure demo}).

Our terracotta warriors data is represented with $\{x_n \in R^p\}_{n=1}^N $, where $R^p$ is the feature space, $x_n$ means the features of one point, such as $x$, $y$, $z$, $N_x$, $N_y$, $N_z$(xyz coordinates and normal value), $N$ means the number of points in one terracotta warrior 3D object.
Our goal is to design a function $f: R^p \rightarrow L$, where $L$ means the segmentation mapping labels, where $\{c_n \in L\}_{n=1}^N$ and $c_n$ is each point's label after the segmentation. In our problem, $\{c_n\}$ is fixed while the mapping function $f$ and $\{x_n\}$ are trainable.
To solve the unsupervised segmentation problem, we can split the problem into two sub-problems. Firstly, we need to design an algorithm to predict optimal $L$. Secondly, we need to design a network to learn the labels $\{c_n\}$ better and make full use of features of terracotta warrior point cloud.
To get the optimal cluster labels $\{c_n\}$, we think that a good point cloud segmentation acts like what a human does. Intuitively, the points with similar features or semantic meanings are easier to segment to the same cluster. In 2D images, spatially continuous pixels tend to have similar colors or textures. In 3D point cloud, grouping spatially continuous points with similar normal value or colors are reasonable to be given the same label in 3D point cloud segmentation. Besides, the Euclidean distance between the points with the same label will not be very long. In conclusion, we design two criteria to predict the $\{c_n\}$:

\begin{itemize}
  \item Points with similar spatial features are desired to be given the same label.
  \item The Euclidean distance between spatially continuous points should not be quite long.
\end{itemize}

We design an SRG algorithm for learning the normal feature of point clouds as much as possible. The experiment \cite{tao2020} shows that the encoder-decoder structure of FoldingNet \cite{yang2017foldingnet} does help neural network learn from point cloud. And DG-CNN is good at learning local features. 
So we propose our SRG-Net inspired by \textbf{FoldingNet} \cite{yang2017foldingnet} and \textbf{DG-CNN} \cite{Wang_2019} to learn part features of terracotta warriors better.
In Section~\ref{Method}, we describe the details of the process of assigning labels $\{c_n\}$. Section~\ref{Experiments} will show that our method achieves better methods compared with other methods.
Compared with existing methods, the key contribution of our work can be summarized as follows:

\begin{itemize}
  \item We design a novel SRG algorithm for point cloud segmentation to make the most of point cloud xyz coordinates by estimating normal features.
  \item We propose our CNN inspired by DG-CNN and FoldingNet to learn the local and global features of 3D point clouds better.
  \item We combine the SRG algorithm and CNN neural network with our refinement method and achieve state-of-the-art segmentation results on terracotta warrior data.
  \item Our end-to-end model not only can be used in terracotta warrior point clouds but also can achieve quite good results on other point clouds. We also evaluate our SRG-Net on ShapeNet dataset.
\end{itemize}

\section{Related Work} \label{Related Work}

Segmentation is typical both in {the} 2D image and 3D point cloud processing.
In image processing, segmentation accomplishes a task {that assigns labels to all the pixels in one image and clusters them with their features}.
Similarly, point cloud segmentation is assigning labels to all the points in a point cloud. The expected result is the points with the similar feature are given with the same label.

In terms of image segmentation, k-means is a classical segmentation method in both 2D and 3D, which leads to partition n observations into k clusters with the nearest mean. {This method is quite popular in data mining}. 
The graph-based method is another popular method like Prim and Kruskal \cite{Kruskal} which implements simple greedy decisions in segmentation. The above methods focus on global features instead of the local differentiated features, so they usually cannot get promising results when it comes to the complex context. 
Among unsupervised deep learning methods, there are many learning features {using the generative methods}, such as \cite{NIPS2009_3674,le2012building,10.1145/1553374.1553453}. They follow the model of neuroscience, where each neuron represents a specific semantic meaning. 
Meanwhile, CNN is widely used in unsupervised image segmentation. For example, in \cite{8462533}, Kanezaki combines the superpixel \cite{Achanta:177415} method and CNN and employs superpixel for back propagation to tune the unsupervised segmentation results. Besides, \cite{Kim_2020} uses a spatial continuity loss as an alternative to settle the limitation of the former work \cite{Achanta:177415}, whose method is also quite valuable for 3D point cloud feature learning.

In the field of 3D point cloud segmentation, the state-of-the-art methods for 2D images are not quite suitable for being directly used on the point cloud. 3D point cloud segmentation methods need to understand both the global feature and geometric details of each point. 
3D point cloud segmentation problems can be classified into semantic segmentation, instance segmentation, and object segmentation. Semantic segmentation focuses on scene-level segmentation instance, segmentation emphasizes object-level segmentation, and object segmentation is centered on part-level segmentation.

As to semantic segmentation, semantic segmentation {aims at} separating a point cloud into several parts with the semantic meaning of each point. 
There are four main paradigms in semantic segmentation, including projection-based methods, discretization-based methods, point-based methods, and hybrid methods. 
Projection-based methods always project a 3D point cloud to 2D images, such as multi-view \cite{lawin2017deep, 10.2312/3dor.20171047}, spherical \cite{DBLP:journals/corr/abs-1710-07368, 8967762}. 
Discretization-based methods usually project a point cloud into a discrete representation, such as volumetric \cite{DBLP:journals/corr/abs-1711-10275} and sparse permutohedral lattices \cite{dai20183dmv, jaritz2019multiview}.
Instead of learning a single feature on 3D scans, several methods {are} trying to learn different parts from 3D scans, such as \cite{dai20183dmv, 8578372_deep_parametric, jaritz2019multiview}.

The point-based network can directly learn features on a point cloud and separate them into several parts. Point clouds are irregular, unordered, and unstructured. 
PointNet \cite{Charles_2017} can directly learn features from the point cloud and retain the point cloud permutation invariance with a symmetric function like maximum function and summation function. PointNet can learn {point-wise} features with the combination of several MLP layers and a max-pooling layer. PointNet is a pioneer that directly learns on the point cloud. A series of point-based networks has been proposed based on PointNet. However, PointNet can only learn features on each point instead of on the local structure. So PointNet++ is presented to get local structure from the neighborhood with a hierarchy network~\cite{qi2017pointnet}. 
PointSIFT \cite{DBLP:journals/corr/abs-1807-00652} is proposed to encode orientation and reach scale awareness. Instead of using K-means to cluster and KNN to generate neighborhoods like the grouping method PointNet++, PointWeb \cite{Zhao_2019_CVPR} {is proposed} to get the relations between all the points constructed in {a local} fully-connected web. As to convolution-based method. 
RS-CNN takes a local point cloud subset as its input and maps the low-level relation to the high-level relation to learn the feature better. 
PointConv \cite{wu2019pointconv} uses the existing algorithm, using a Monte Carlo estimation to define the convolution. 
{PointCNN \cite{li2018pointcnn} uses $\chi-conv$ transformation to convert the point cloud into a latent and canonical order.} As to point convolution methods, Parametric Continuous Convolutional Neural Network(PCCN) \cite{8578372} is proposed based on parametric continuous convolution layers, whose kernel function is parameterized by MLPs and spans the continuous vector space. 
Graph-based methods can better learn the features like shapes and geometric structures in point clouds. Graph Attention Convolution(GAC)~\cite{2018graph} can learn several relevant features from local {neighborhoods} by dynamically assigning attention weights to points in different {neighborhoods} and feature channels. 
Dynamic Graph CNN(DG-CNN) \cite{Wang_2019} constructs several dynamic graphs in {neighborhood}, and concatenates the local and global {features} to extract better features and update them each graph after each layer of the network dynamically. 
FoldingNet {adopts} the auto-encoder structure to encode the point cloud $N\times3$ to $1\times512$ and decode it to $M\times3$ with the aid of chamfer loss to construct the auto-encoder network.

Part segmentation is more complex than semantic and instance segmentation because {there are significant geometric differences between the points with the same labels}, and the number of parts with the same semantic meanings may differ. 
{Z. Wang et al.} \cite{wang2019voxsegnet} propose VoxSegNet to achieve promising part segmentation results on 3D voxelized data, which presents a Spatial Dense Extraction(SDE) module to extract multi-scale features from volumetric data. 
Synchronized Spectral CNN (SyncSpecCNN) \cite{yi2016syncspeccnn} is proposed to achieve fine-grained part segmentation on irregularity and non-isomorphic shape graphs with convolution. 
\cite{LIU2008576} is proposed to {segment unorganized noisy point clouds automatically} by extracting clusters of points on the Gaussian sphere. 
\cite{DIANGELO201544} uses three shape indexes: {the smoothness indicator, shape index, and flatness index based on a fuzzy parameterization}. \cite{BENKO2004511} presents a segmentation method for conventional engineering objects based on local estimation of various geometric features. 
Branched AutoEncoder network (BAE-NET) \cite{chen2019baenet} is proposed to perform unsupervised and {weakly-supervised} 3D shape co-segmentation. Each branch of the network can learn features from a specific part shape for a particular part shape with representation based on the auto-encoder structure.

\section{Proposed Method} 
\label{Method}

In this paper, our input data for the terracotta warrior is {in the form of} 3D point clouds (see samples in Fig.~\ref{figure demo}). Point cloud data is represented as a set of 3D points $\{P_{i} | i=1, 2, 3...n\}$, where each point is a vector $R^n$ containing coordinates $x$, $y$, $z$ and other features like normal, color. Our method contains three steps:

\begin{enumerate}
  \item We compute normal value with the $xyz$ coordinates.
  \item We use our seed-region-growing (SRG) method to pre-segment the point cloud.
  \item We use our pointwise CNN called SRG-Net self-trained to segment the point cloud with the refinement of pre-segmentation results.
\end{enumerate}

Given the point cloud only with 3D coordinates, there are several effective normal estimation methods like \cite{OUYANG20051071} \cite{ZHOU2020102916} \cite{WANG20131333}. In our method, we follow the simplest method in \cite{Normal_Estimation} because of its low average-case complexity and quite high accuracy. We propose an unsupervised segmentation method for the terracotta warrior point cloud by combining our seed-region-growing clustering method with our 3D pointwise CNN called SRG-Net. This method contains two phases, SRG segmentation and segmentation network. The problem we want to solve can be described as follows. Firstly, the SRG algorithm is implemented on the point cloud to do pre-segmentation. Secondly, we use SRG-Net for unsupervised segmentation with pre-segmentation labels.

\subsection{Seed Region Growing} \label{Seed Region Growing}

Unlike 2D images, not all point cloud data has features such as color and normal. For example, our terracotta warrior 3D objects do not have any color feature. However, normal vectors can be calculated and predicted by point coordinates \cite{WANG20131333}. It is worth noting that there are many similarities and differences between color in 2D and normal features in 3D. For color features in a 2D image, if the pixels are semantically related, the color of the pixel in the neighborhood generally does not change a lot. For 3D point cloud normal features, compared with the color features in 2D images, the normal values of points in the neighborhood of the point cloud often differ. 
Points of neighborhood in one point cloud share similar features.

We design a seed-region-growing method to pre-segment the point cloud based on the above characteristics of the normal feature of the point cloud.


First, we implement KNN to the point cloud to get the nearest neighbors of each point. Then we initialize a random point as the start seed and add to the available points to start the algorithm. Then we choose the first seed from the available list to judge the points in its neighborhood. If the normal value and Euclidean distance are within the threshold we set, we think that the two points are semantically continuous, and we can group two points into one cluster. The outline about the SRG is given in { \textbf{Algorithm \ref{Algorithm1}}.} The description of the algorithm is as follows: 

\begin{enumerate}
  \item Input:
  \begin{enumerate}
    \item xyz coordinates of each point ${p_n \in R^3}$ 
    \item normal value of each point ${n_n \in R^3}$
    \item Euclidean distance of every two points ${e_n \in R^n}$
    \item neighborhood of each point ${b_n \in R^k}$
  \end{enumerate}
  
  \item The randomly chosen point is added to the set $A$, we can call it seed $a$.
  \item For other points which are not transformed to seeds in $A$:
  \begin{enumerate}
    \item Each of point in seed's {neighbor} $N_a$, check the normal distance and Euclidean distance between the point in $N_a$ and the seed $a$.
    \item We set the normal distance and the Euclidean distance threshold. {If both of the two value is less than the two thresholds, we can add it to $A$.}
    \item The current seed is added to the checked list $S$.
  \end{enumerate}
\end{enumerate}

\newcommand\mycommfont[1]{\footnotesize\ttfamily\textcolor{blue}{#1}}
\SetCommentSty{mycommfont}

\SetKwInput{KwInput}{Input}                
\SetKwInput{KwOutput}{Output}              
\SetKwInput{KwData}{Initialize}              

\begin{algorithm} 
\DontPrintSemicolon
  
  \KwInput{ $P = \{p_n\in R^3\}$ \tcp*{x, y, z coordinates} \\
    $N = \{n_n\in R^3\}$ \tcp*{normal value} \\
    $E = \{e_n\in R^n\}$ \tcp*{Euclidean distance} \\
    $\Omega(\cdot)$ \tcp*{nearest neighbour function} \\
  }
  \KwOutput{$L = \{l_n\in R^1\}$ \tcp*{segmentation label of each point}}
  \KwData{$S := \diameter$ \tcp*{set seed list to empty} \\
    $A := rand \{1, 2... |P|\}$ \tcp*{select random point to available point list as seed}
  }
  \For{t = 1 to T}
  {
    \While{$A \ne \diameter$}
    {
   	  $a \leftarrow first\; item\; in\; A$ \\
   	  $neighbours \leftarrow \Omega(a)$
   	  \For{$neighbours \ne \diameter$}
      {
        $neighbour \leftarrow rand(neighbours)$\\
        \If{$E(a, neighbour) \le e_th \land N(a, neighbour) \le e_th 
          \land neighbour \not\in S $ } 
        {
          $append\; neighbour\; to\; S$
          $remove\; neighbour\; from\; neighbours$
        }
        \Else{
          $remove\; neighbour\; from\; neighbours$
         }
      }
      $add\; a\; to\; S$  
    }
  }
\caption{SRG unsupervised segmentation} \label{Algorithm1}
\end{algorithm}

\subsection{SRG-Net} \label{SRG-Net}

\begin{figure*}
	\includegraphics[width=\textwidth]{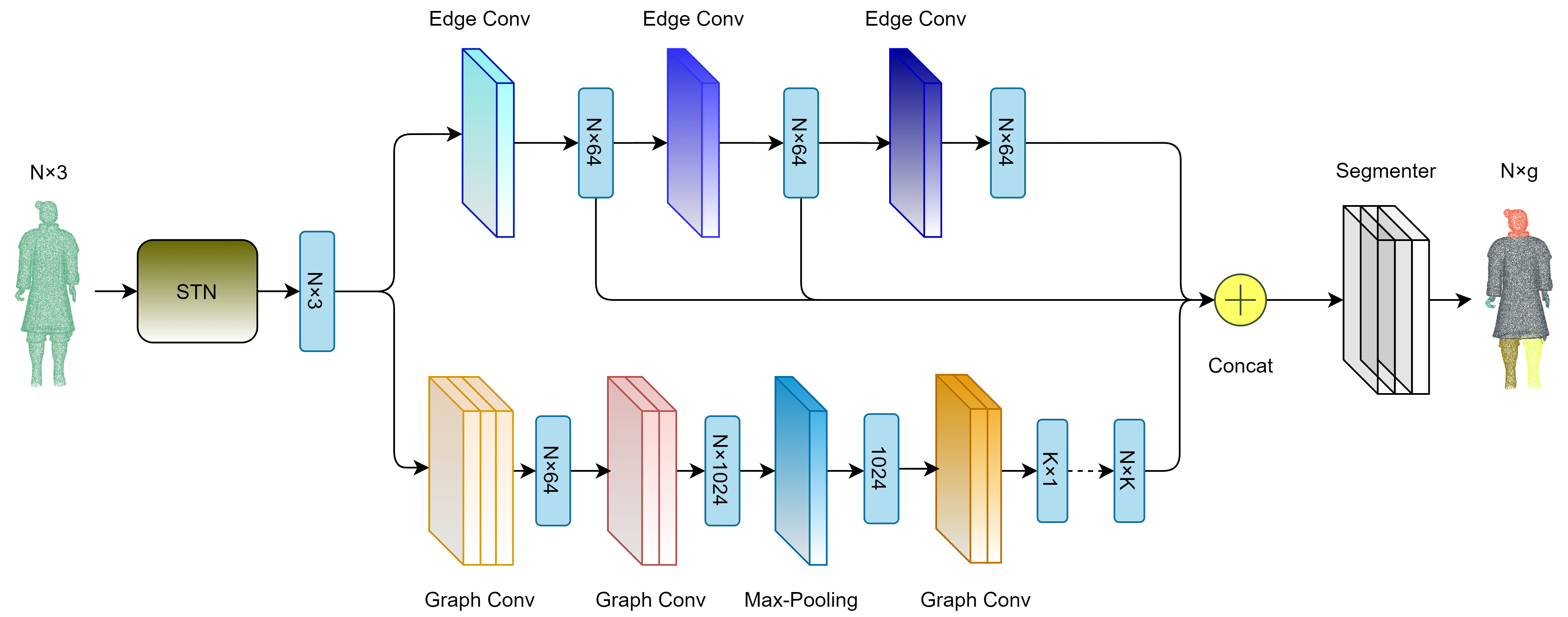}
	\caption{Network Structure. {The brown part represents the Spatial Transformer Network, the blue module represents the Edge Convolution, and the lower module represents the Graph Convolution. The tensors of the upper and lower modules will be concatenated, and be input into the Segmenter to get the final segmentation result. (Segmenter will be explained in the following figure.)}}\label{figure 1}
	\label{network structure}
\end{figure*}

We propose our SRG-Net, which is inspired by the dynamic graph in DG-CNN and auto-encoder in FoldingNet. Unlike classical graph CNN, our graph is dynamic and updated in each layer of the network. Compared with the methods that only focus on the relationship between points, we also propose an encoder structure to better express the features of the entire point cloud, aiming at learning the structure of the point cloud and optimize the pre-segmentation results of SRG. Our network structure can be shown in Fig.~\ref{figure 1}, which consists of two sub-network, the first part is an encoder that generates features from the dynamic graph and the whole point cloud and the second is the segmentation network.

We denote the point cloud as $S$. We use lower-case letters to represent vectors, such as $x$, and use upper-case letter to represent matrix, such as A. We call a matrix m-by-n or $m \times n$ if it has m rows and n columns. In addition, the terracotta warrior point cloud data is N points with 6 features $x$, $y$, $z$, $N_x$, $N_y$, $N_z$(xyz coordinates and normal values), denoted as $X = \{x_1, x_2, x_3,  ...    , x_n\} \subseteq R^6$.

\subsubsection{Encoder Architecture}

The SRG-Net encoder follows a similar design of \cite{yang2017foldingnet}, the structure of SRG-Net is shown in Fig.~\ref{network structure}. Compared with \cite{yang2017foldingnet}, our encoder concatenate several multi-layer perceptrons(MLP) and several dynamic graph-based max-pooling layers. The dynamic graphs are constructed by applying KNN on point clouds. 


For the entire point cloud, we compute a spatial transformer network and get a transformer matrix of 3-by-3 to maintain invariance under transformations. Then for the transformed point cloud, we compute three dynamic graphs and get graph features respectively. In graph feature extracting process, we adopt the Edge Convolution in \cite{le2012building} to compute the graph feature of each layer, {which uses an asymmetric edge function in Eq.~(\ref{eq:1}):
\begin{equation} \label{eq:1}
{f_i}_j = h(x_i, x_i-x_j)  
\end{equation}
where it combines the coordinates of {neighborhood} center $x_i$ with the subtraction of neighborhood point and the center point coordinates $x_i-x_j$ to get local and global information of neighborhood.} Then we define our operation in {Eq.~(\ref{eq:2}): 
\begin{equation} \label{eq:2}
g_{ij} = \Theta(\mu \cdot (x_i - x_j) + \omega \cdot x_i)
\end{equation}
where $\mu$ and $\omega$ are parameters and $\Theta$ is a ReLU function. Eq.~(\ref{eq:2}) is implemented as a shared MLP with Leaky ReLU.}
Then we define our max-pooling operation in {Eq.~(\ref{eq:3}): 
\begin{equation} \label{eq:3}
x_i = \max_{j \in N(i)} g_{ij}
\end{equation}
where $N(i)$ means neighborhood of point $i$.}

The bottleneck is computed by the graph feature extraction layer. The structure is shown in Fig.~\ref{network structure}. First, we compute the covariance $3 \times 3$ matrix for every point and vectorize it to $1 \times 9$. Then the $n \times 3$ matrix of point coordinates is concatenated with the $n \times 9$ covariance matrix into a $n \times 12$ matrix. Then we put the matrix into a 3-layer perceptron. Then we feed the output of the perceptron to two subsequent graph layers. In each layer, max-pooling is added to the neighbor of each node. At last, we apply a 3-layer perceptron to the former output and get the final output. The whole process of the graph feature extraction layer is summarized in Eq.~(\ref{eq:4}):

\begin{equation} \label{eq:4}
Y = I_{max} (X) K
\end{equation}
In {Eq.~(\ref{eq:4}), $X$ is the input matrix to the graph layer and $K$ is a feature mapping matrix. $I_{max} (X)$ can be represented in Eq. \ref{eq:5}:
\begin{equation} \label{eq:5}
(I_{max} (X))_{ij} = \Theta(\max_{k\in N(i)}x_{kj})
\end{equation}
where $\Theta$ is a ReLU function and $N(i)$ is the {neighborhood} of point $i$.}
The max-pooling operation in {Eq.~(\ref{eq:5})} can get local feature based on the graph structure. So the graph feature extraction layer can not only get local neighborhood features, but also global features. 

\begin{figure}
    \centering
	\includegraphics[width=0.4\textwidth]{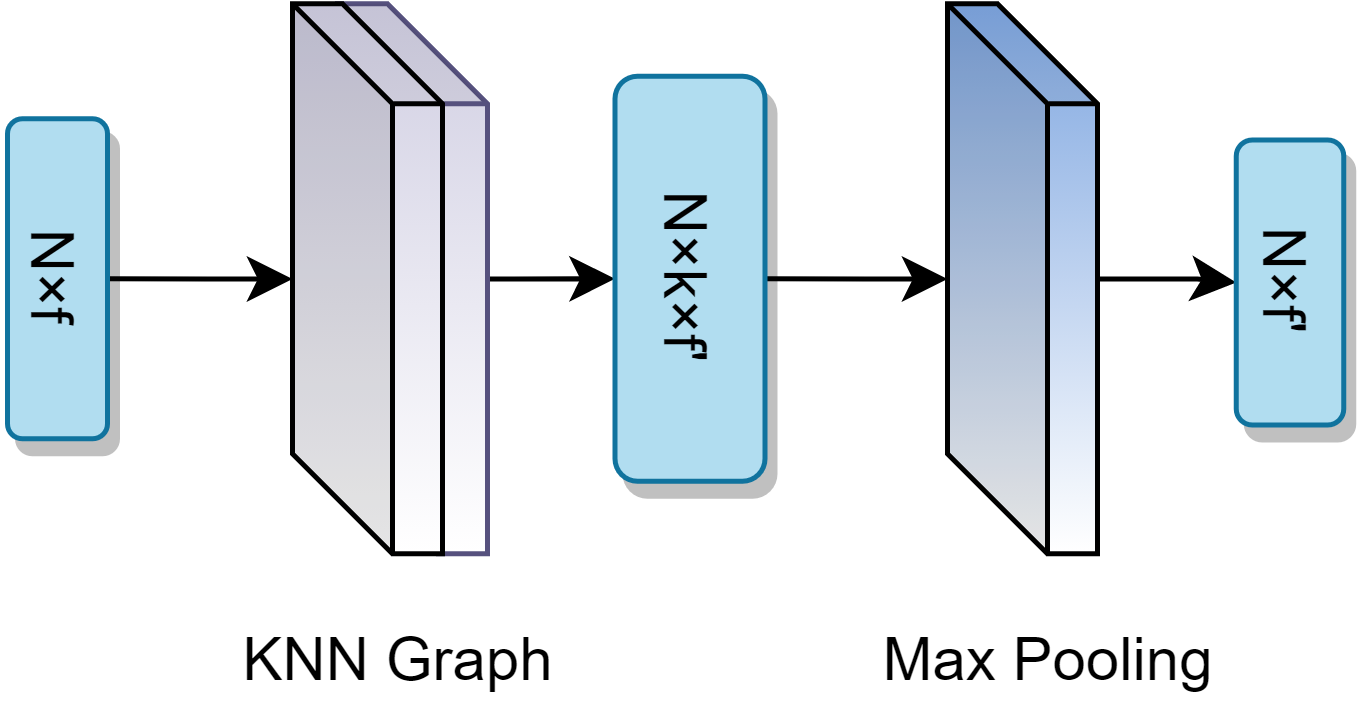}
	\caption{Edge Convolution. {The figure contains a KNN graph convolution mapping the point cloud to N × k × f', and then put into max-pooling to get the point cloud edge feature.}}
	\label{edge convolution}
\end{figure}

\subsubsection{Segmenter Architecture}

\begin{figure*}[htbp]
	\includegraphics[width=\textwidth]{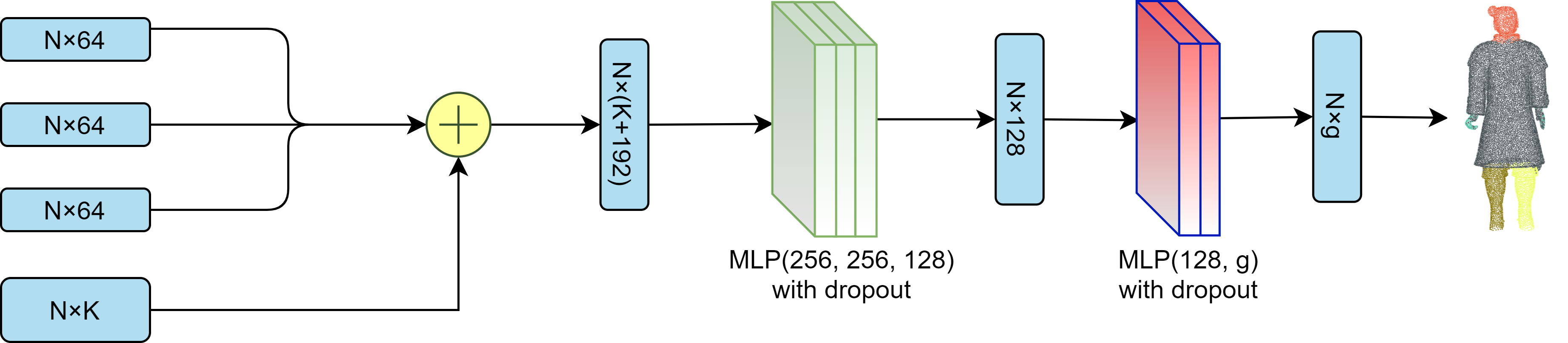}
	\caption{Segmenter Structure. {First, the segmenter concatenates the features calculated by edge convolution and graph convolution. Then put into three-layer-perceptron, and finally put into two-layer-perceptron to get the final segmentation result.}}
	\label{segmenter structure}
\end{figure*}

Segmenter gets dynamic graph features and bottleneck as input and assign labels to each point to segment the whole point cloud. The structure of segmenter is shown in Fig.~\ref{network structure}. {First, bottleneck is replicated $N$ times in {Eq.~(\ref{eq:6})}: 
\begin{equation} \label{eq:6}
B' = \underbrace{B \cup B  \cup ... \cup B}_{N}
\end{equation}
where $N$ is the number of points in point cloud and $B$ is the bottleneck.}
The output of replication is concatenated with dynamic features in Eq.~(\ref{eq:7}): 
\begin{equation} \label{eq:7}
C = B' \cup D_1 \cup D_2 \cup D_3 
\end{equation}
where $D_1$, $D_2$, $D_3$ represent dynamic graph features.
At last, we feed the output of concatenation to a multi-layer-perceptron to segment the point cloud in Eq.~(\ref{eq:8}). 
\begin{equation} \label{eq:8}
D = \Theta(\Psi \cdot C + \Omega) 
\end{equation}
where $\Psi$ and $\Omega$ represent parameters in the linear function, and $\Theta$ represents a ReLU function.

\subsection{Refinement} \label{refinement}

Inspired by \cite{Kanezaki}, we designed point refinement to achieve better segmentation results. 
The basic concept of point clustering is to group similar points into clusters. In point cloud segmentation, it is intuitive for the clusters of points to be spatially continuous. We add the constraint to favor cluster labels that are the same as those of neighboring points. We first extract $K$ superpoints $\{S_k\}_{k=1}^K$ from the input point cloud $L = \{v_n\}_{n=1}^N$, where $S_k$ is a set of the indices of points that belongs to the k-th superpoint. Then we assign points in each superpoint to have the same cluster label.
More specifically, letting $|c_n|_{n \in S_k}$ be the number of points in $S_k$ that belong to the $c_n$th cluster, we select the most frequent cluster label $c_max$, where $|c_{max}|_{n \in S_k} \geq |c_n|_{n \in S_k}$ for all $c_n in {1, ..., q}$. The cluster labels are replaced by $c_{max}$ for $n \in S_k$.

In this paper, there is no need to set a large number in seed-region-growing process because our neural network can learn both the local and global features. We choose seed-region-growing method with $K = 25$ for the super point extraction. The refinement process can achieve better results compared with straight learning in Fig.~\ref{different methods figure}. In addition, instead of using ground truth to calculate loss, we use the coarse segmentation result of seed-region-growing method to calculate the loss.

\section{Experiments} \label{Experiments}

We conduct experiments on terracotta warrior dataset and ShapeNet dataset. We implement the pipeline using PyTorch and Python3.7. All the results are based on experiments under RTX 2080 Ti and i9-9900K. The performances of each method in the experiment are evaluated by the accuracy (mIoU) and the latency.

\subsection{Experiments on Terracotta Warrior}

\newcolumntype{P}[1]{>{\centering\arraybackslash}p{#1}}

We use Artec Eva \cite{artec} to collect 500 intact terracotta warrior models, and we take 400 of the 500 models as the training set and 100 as the validation set. Each model consists of about 2 million points which include xyz coordinates, vertical normals, triangle meshes and RGB data. Before the experiments start, we eliminate the triangle meshes and RGB data of the original models, and remain xyz coordinates and vertical normals. Moreover, we uniformly sample the above point clouds to 10,000 points thus as the experimental inputting. In reality, the terracotta warriors are generally unearthed in the form of limb fragments. 

To better assist the restoration of terracotta warriors, we split the point cloud into six pieces both in SRG and K-Means. We also set the number of iterations $T$ as 2000 in each method(PointNet\cite{Charles_2017}, PointNet2\cite{qi2017pointnet}, DG-CNN\cite{Wang_2019}, PointHop++\cite{zhang2020pointhop++}, ECC\cite{Simonovsky2017ecc}) for every point cloud. Unlike the supervised problem, our unsupervised method solves two sub-problems: prediction of cluster labels with fixed network parameters and training of network with predicted labels. The former sub-problem is solved with Section~\ref{refinement}. The latter sub-problem is solved with back-propagation. 

We select SGD as optimizer, choose $lr = 0.005$, and set the momentum factor to 0.1. To make fair comparison in the experiments, we combine PointNet\cite{Charles_2017}, PointNet2\cite{qi2017pointnet}, DG-CNN\cite{Wang_2019}, PointHop++\cite{zhang2020pointhop++} with the seed region growing method to have a similar structure of SRG-Net. In addition, the seed region growing is also compared with K-means.
Comparison of different methods is shown in Table~\ref{comparison}, and the visualization results of different methods are presented in Table~\ref{different methods figure}. As is shown in Table~\ref{different methods figure}, compared with k-means methods(K-means-DGCNN, K-means-Pointnet2, K-means-Pointnet), our seed-region-growing method can get roughly correct segmentation results. In contrast, k-means methods get the wrong segmentation results. Compared with SRG-DGCNN, SRG-PointNet2, SRG-PointNet, our SRG-Net can get more accurate segmentation results.


As shown in Table~\ref{comparison}, SRG-Net outperforms SRG-DGCNN by increasing more than 5.5\% of accuracy with consuming about 50\% of SRG-DGCNN latency. Our method also outperforms SRG-PointHop++ and SRG-ECC by a large margin(6.1\%, 8.2\%) with about 72.2\% and 70.3\% latency. Compared with K-means methods, SRG-Net improves SRG-ECC and SRG-DGCNN by more than 10\% in mIoU, and it also outperforms Kmeans-PointNet2 and K-means\&PointNet by a large margin.

\begin{table}[htbp]\footnotesize
    \center
	\begin{tabular}{ P{0.32\textwidth} P{0.24\textwidth} P{0.24\textwidth}} \toprule
		Method & Accuracy(\%) & Latency(ms) \\
		\midrule
		SRG-DGCNN & 78.32 & 37.26 \\
		SRG-PointHop++ & 77.94 & 30.62 \\
		SRG-ECC & 76.32 & 31.45 \\
		SRG-Pointnet & 72.03 & 13.35 \\
		SRG-Pointnet2 & 70.55 & 24.10 \\
		\midrule
		K-means-DGCNN & 70.64 & 36.30 \\
		K-means-Pointnet2 & 52.33 & 25.38 \\
		K-means-Pointnet & 53.94 & 13.47 \\
		\midrule
	    SRG-Net & \textbf{82.63} & \textbf{22.11} \\
		\bottomrule
	\end{tabular}
	\caption{Comparison of Different Methods} \label{comparison}
\end{table}

Ranking the results of different methods in Table~\ref{comparison}, we can draw the following conclusions: 
\begin{enumerate}
    \item Compared with SRG-DGCNN, SRG-PointNet, SRG-PointNet2, our network can enhance the accuracy of learning with less latency.
    \item Compared with Kmeans-DGCNN, Kmeans-PointNet, Kmeans-PointNet2, our customized SRG method can achieve better results compared with k-means.
    \item In general, SRG-Net has obvious advantages in accuracy and latency.
\end{enumerate}
 
\newcommand\x{2.5cm}
\begin{table*}[htbp]\centering \label{table 2}\footnotesize
	\begin{tabular}{ >{\centering\arraybackslash} m{0.1\textwidth} >{\centering\arraybackslash}    >{\centering\arraybackslash} m{0.12\textwidth} >{\centering\arraybackslash} m{0.12\textwidth} >{\centering\arraybackslash} m{0.12\textwidth} >{\centering\arraybackslash} m{0.12\textwidth} >{\centering\arraybackslash} m{0.12\textwidth} >{\centering\arraybackslash} m{0.12\textwidth} }  \toprule
		Method & 005413 & 005420 & 005422 & 005423 & 005427 & 005455\\
		\midrule
		SRG-DGCNN &
		\includegraphics[height=\x,keepaspectratio]{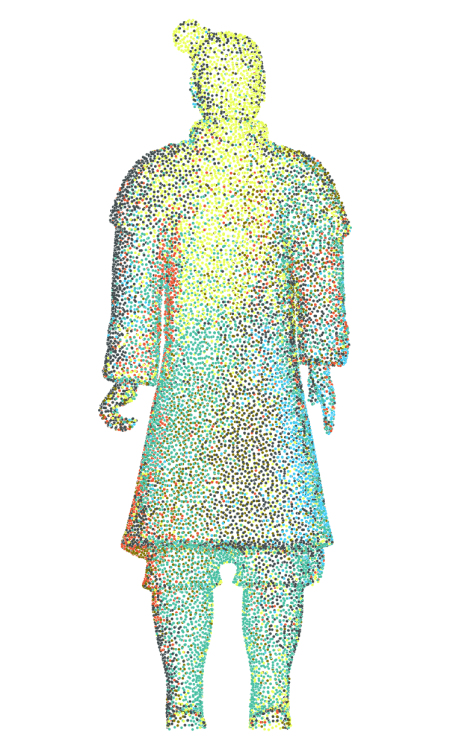} &
		\includegraphics[height=\x,keepaspectratio]{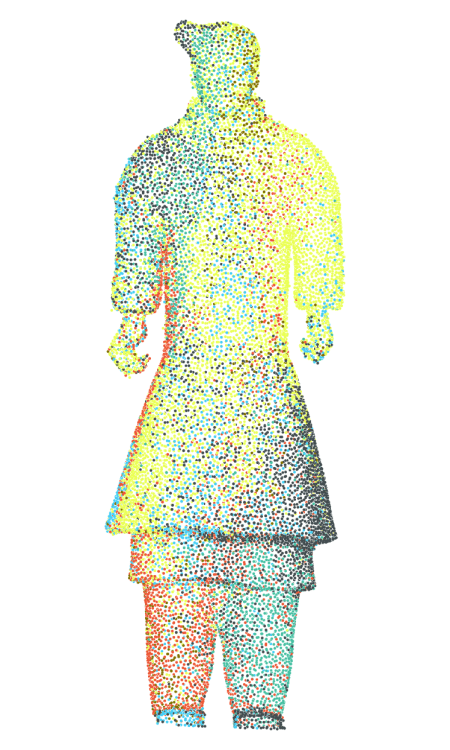} &
		\includegraphics[height=\x,keepaspectratio]{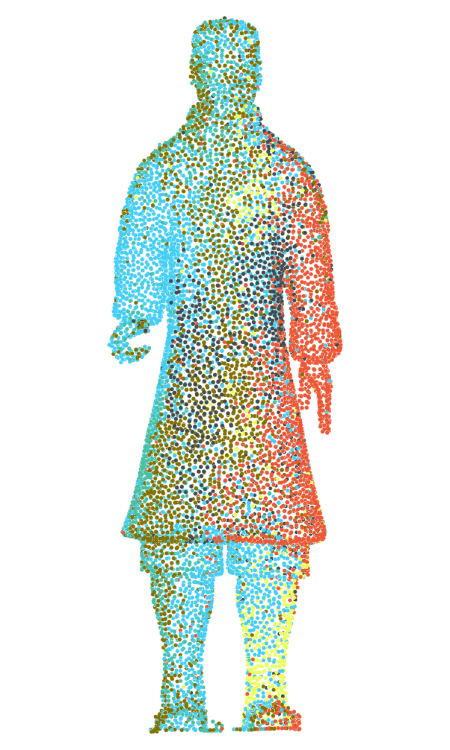} &
		\includegraphics[height=\x,keepaspectratio]{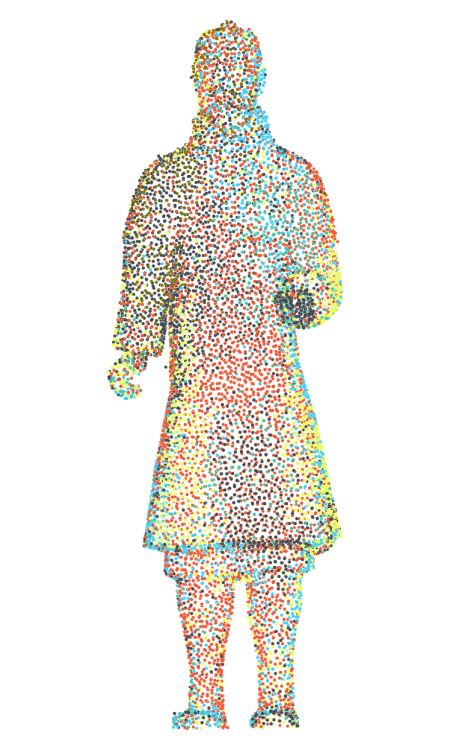} &
		\includegraphics[height=\x,keepaspectratio]{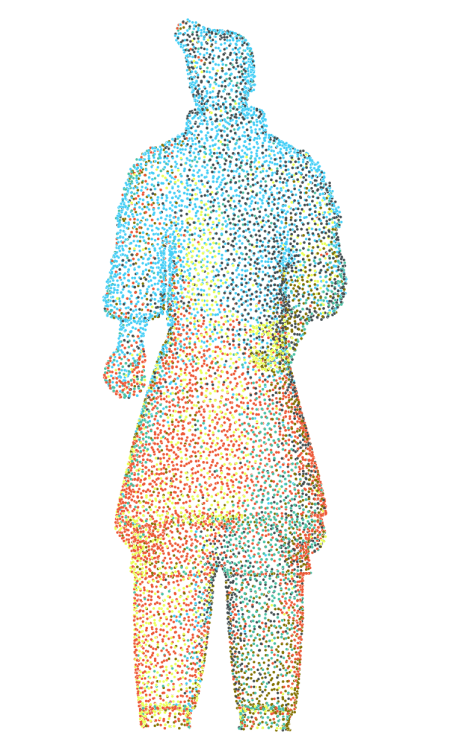} &
		\includegraphics[height=\x,keepaspectratio]{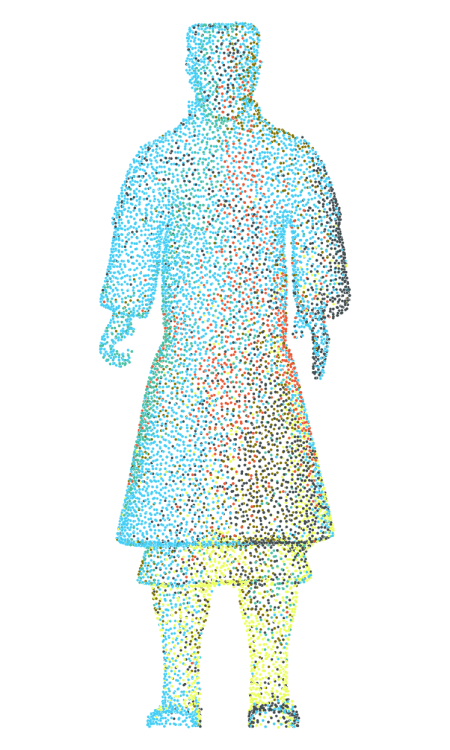} \\
		SRG-Pointnet2 &
		\includegraphics[height=\x,keepaspectratio]{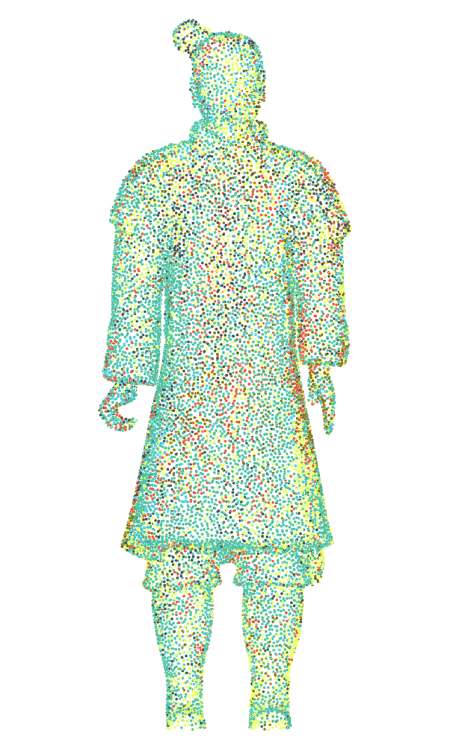} &
		\includegraphics[height=\x,keepaspectratio]{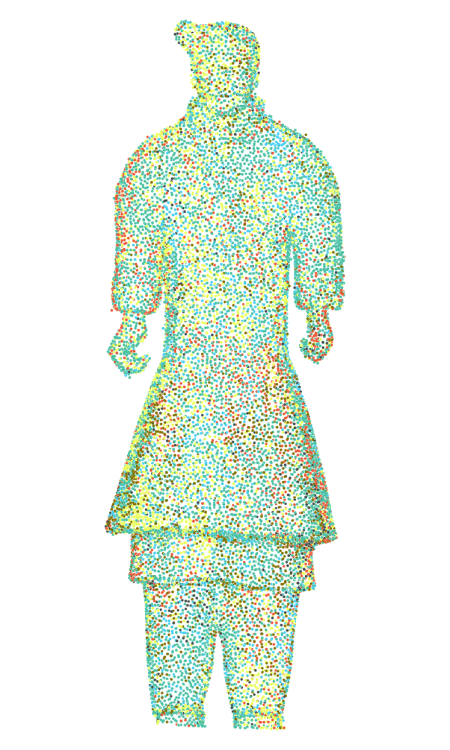} &
		\includegraphics[height=\x,keepaspectratio]{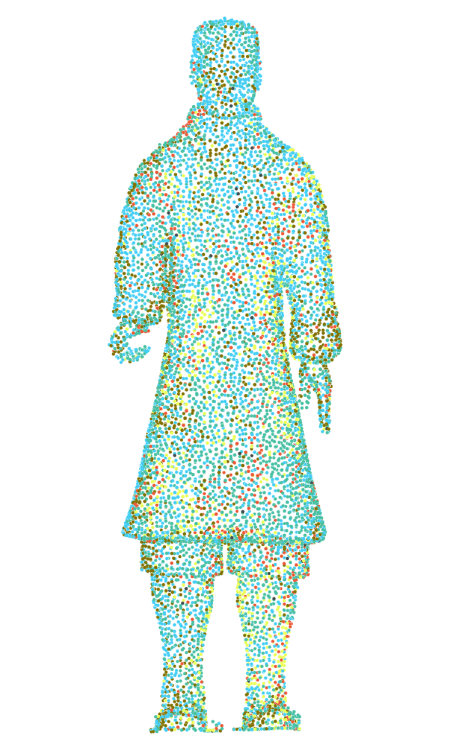} &
		\includegraphics[height=\x,keepaspectratio]{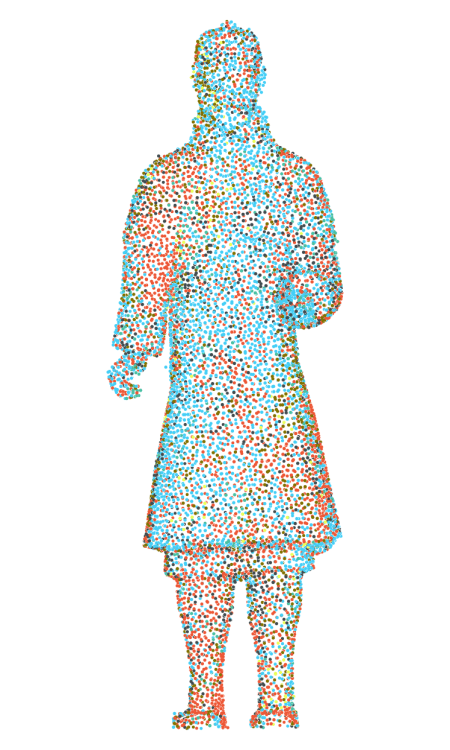} & 
		\includegraphics[height=\x,keepaspectratio]{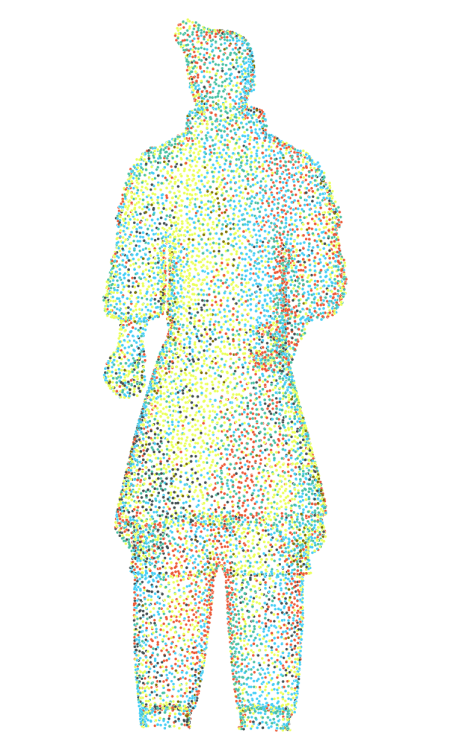} &  
		\includegraphics[height=\x,keepaspectratio]{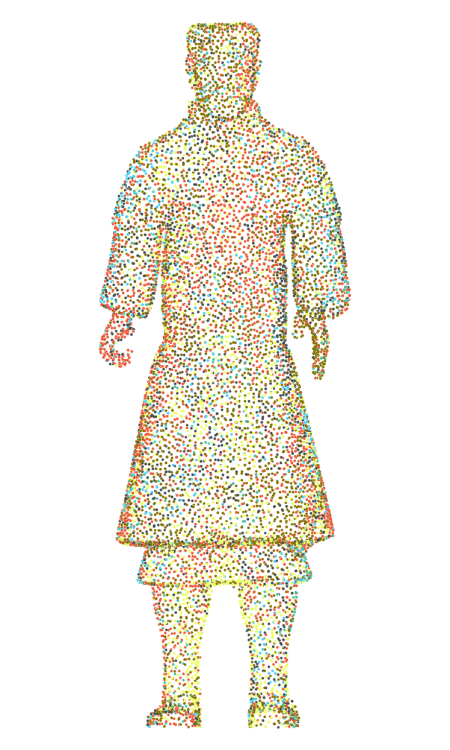}   \\
		SRG-Pointnet &
		\includegraphics[height=\x,keepaspectratio]{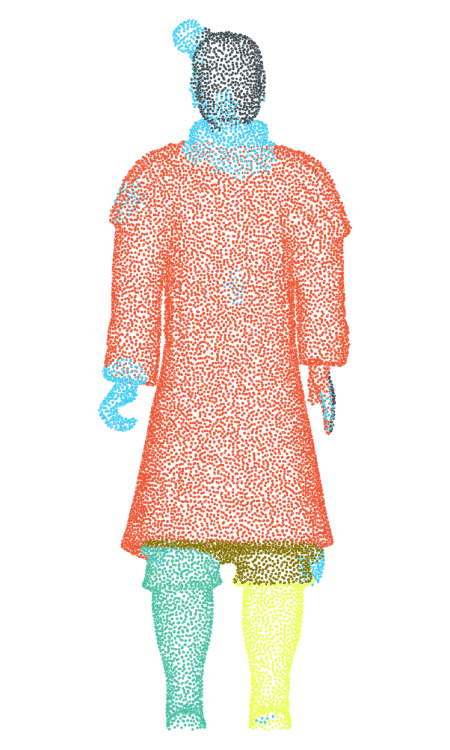} &
		\includegraphics[height=\x,keepaspectratio]{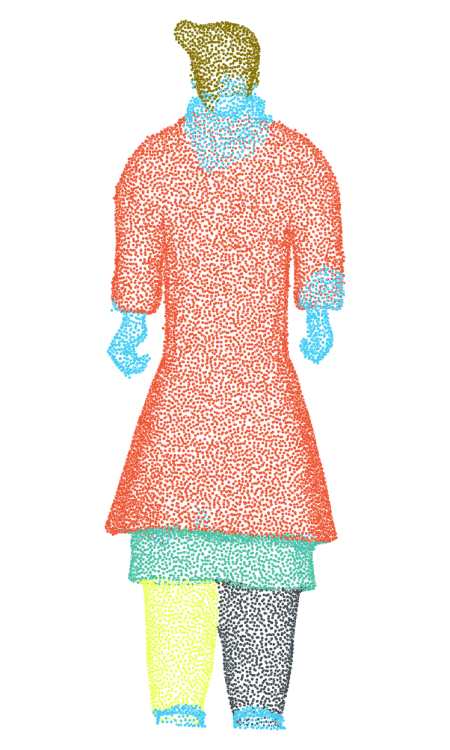} &
		\includegraphics[height=\x,keepaspectratio]{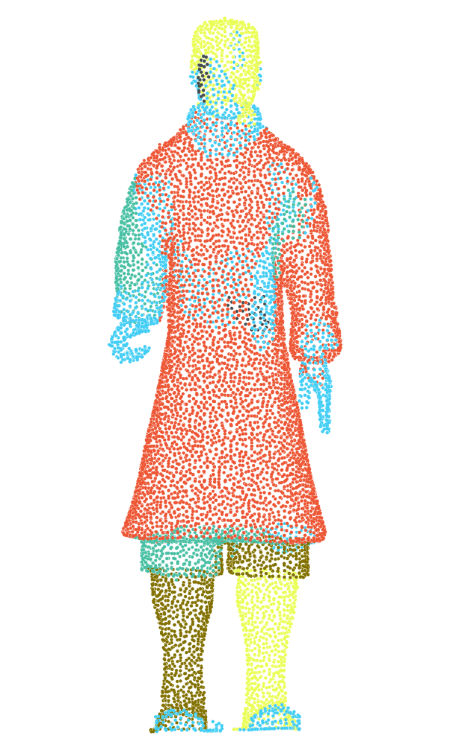} &
		\includegraphics[height=\x,keepaspectratio]{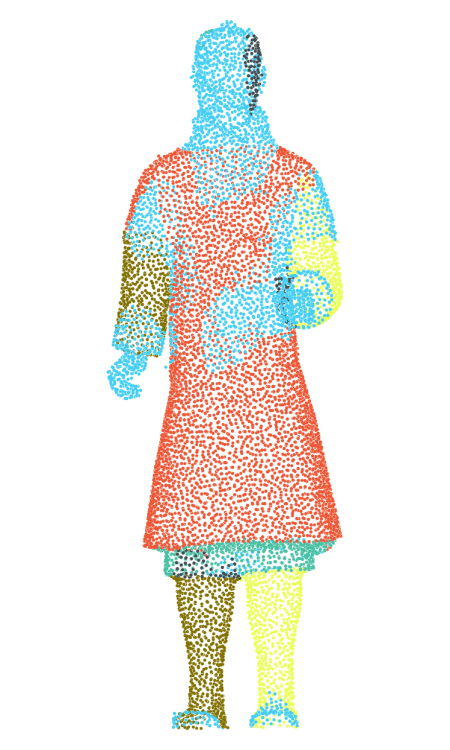} &
		\includegraphics[height=\x,keepaspectratio]{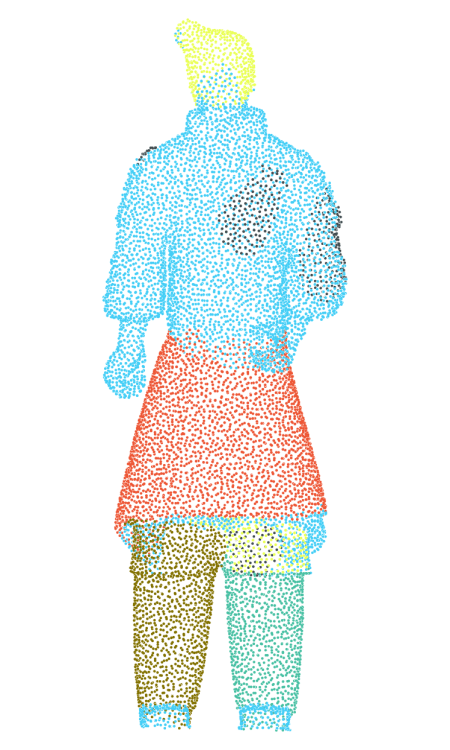} &
		\includegraphics[height=\x,keepaspectratio]{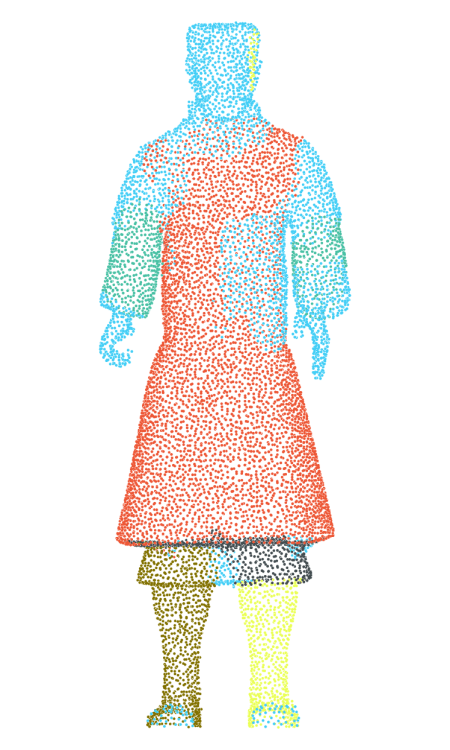}   \\
		\midrule
		K-means-DGCNN &
		\includegraphics[height=\x,keepaspectratio]{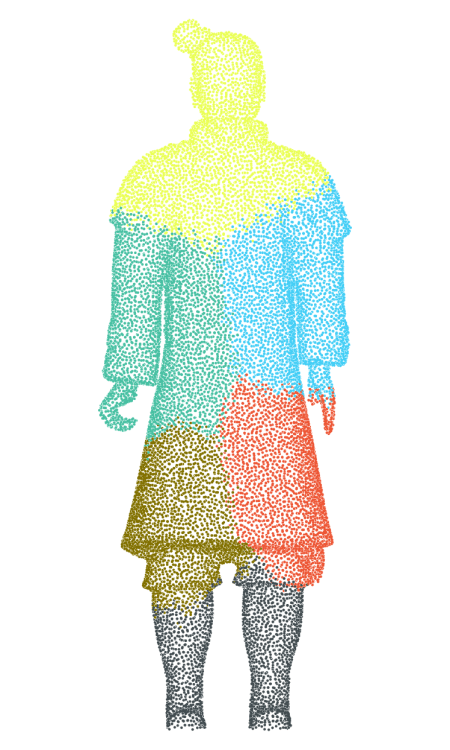} &
		\includegraphics[height=\x,keepaspectratio]{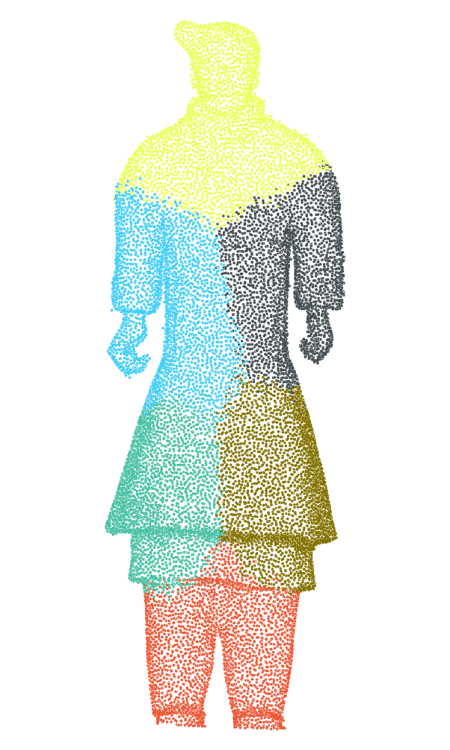} &
		\includegraphics[height=\x,keepaspectratio]{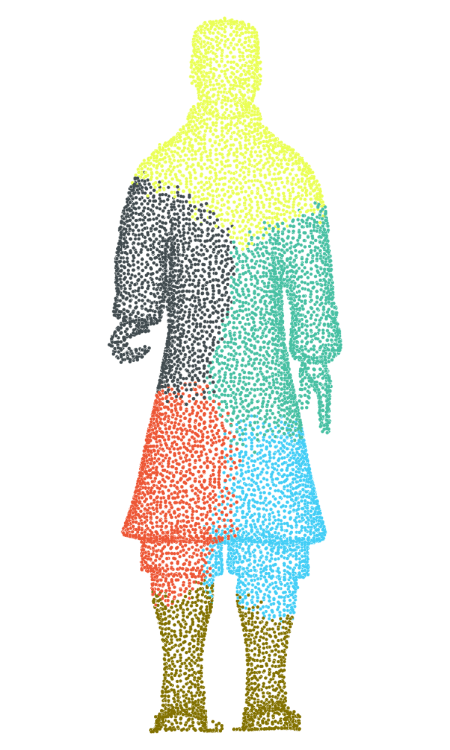} &
		\includegraphics[height=\x,keepaspectratio]{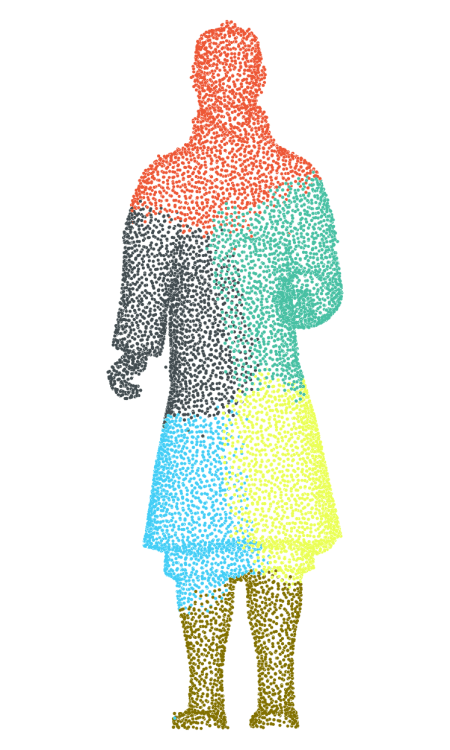} &
		\includegraphics[height=\x,keepaspectratio]{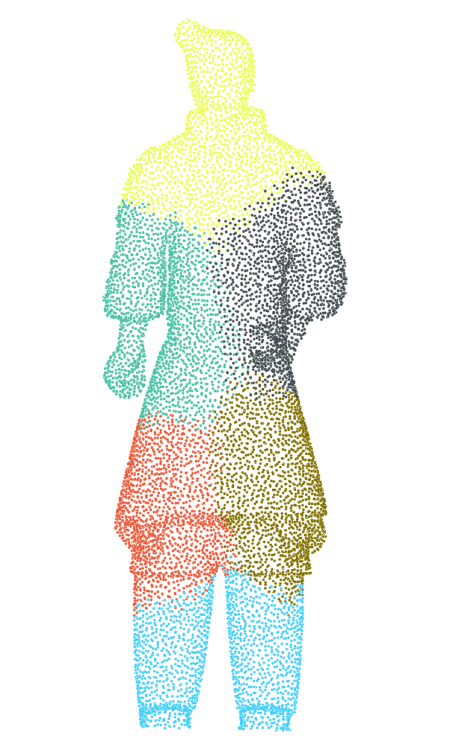} &
		\includegraphics[height=\x,keepaspectratio]{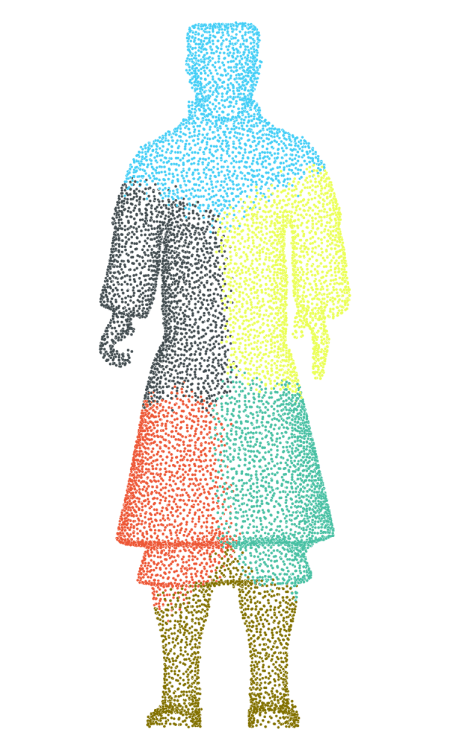}   \\
		K-means-Pointnet2 &
		\includegraphics[height=\x,keepaspectratio]{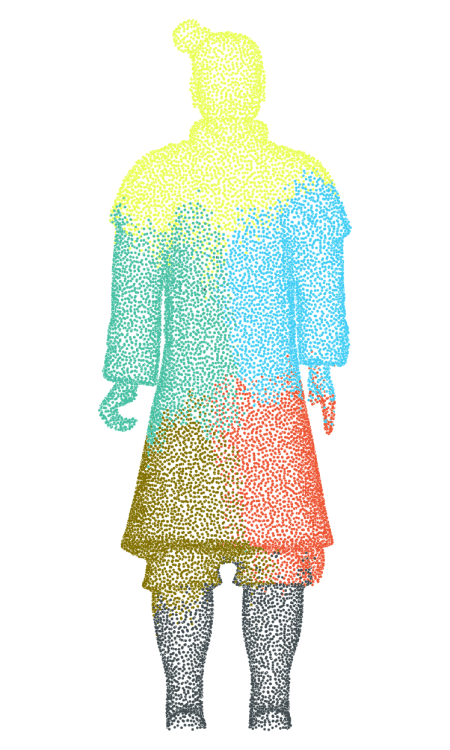} &
		\includegraphics[height=\x,keepaspectratio]{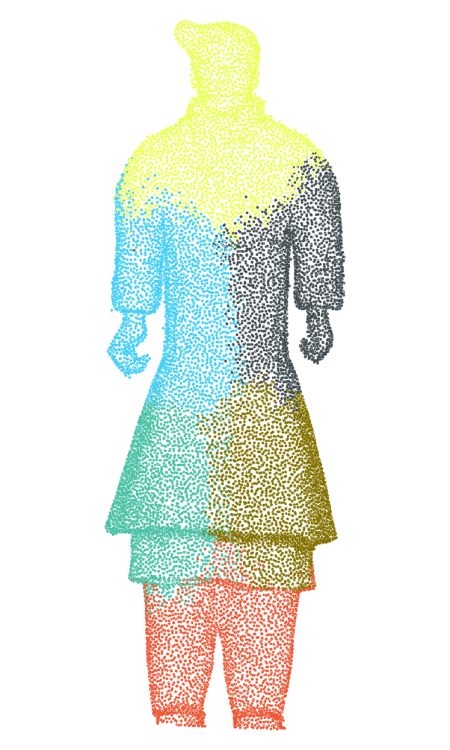} &
		\includegraphics[height=\x,keepaspectratio]{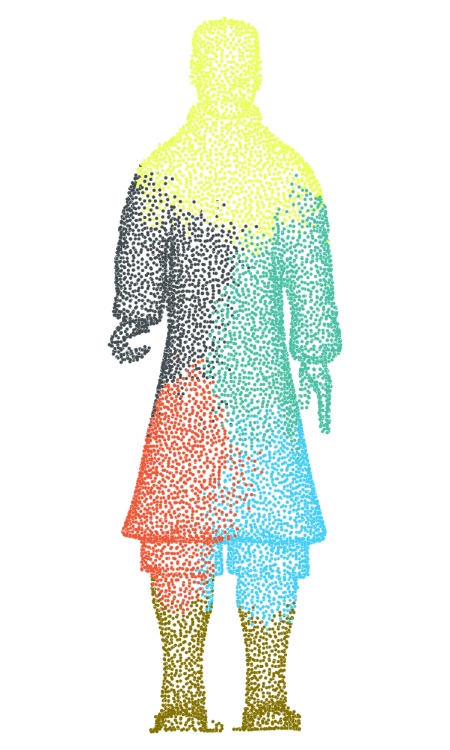} &
		\includegraphics[height=\x,keepaspectratio]{pictures/result/005422/refine_kmeans_pointnet2.png} &
		\includegraphics[height=\x,keepaspectratio]{pictures/result/005422/refine_kmeans_pointnet2.png} &
		\includegraphics[height=\x,keepaspectratio]{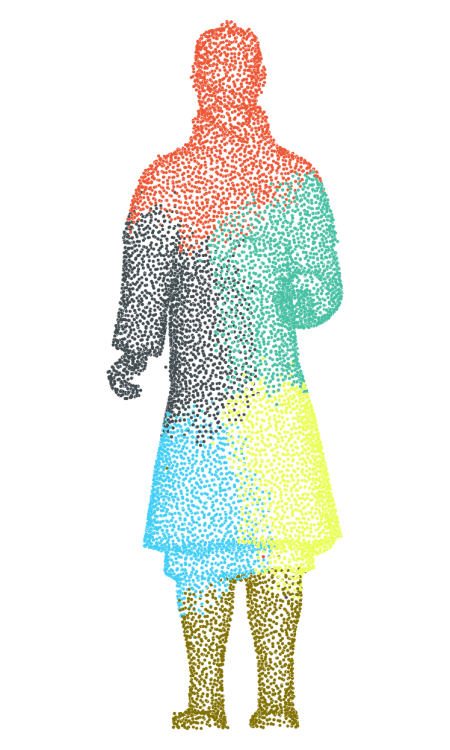}   \\
		Kmeans-Pointnet &
		\includegraphics[height=\x,keepaspectratio]{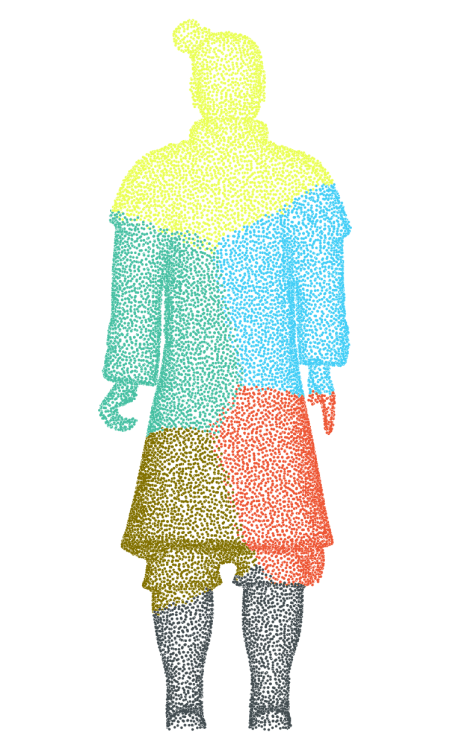} &
		\includegraphics[height=\x,keepaspectratio]{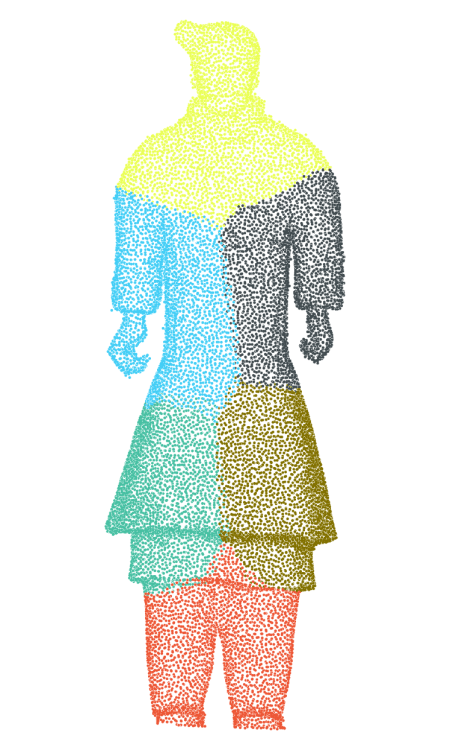} &
		\includegraphics[height=\x,keepaspectratio]{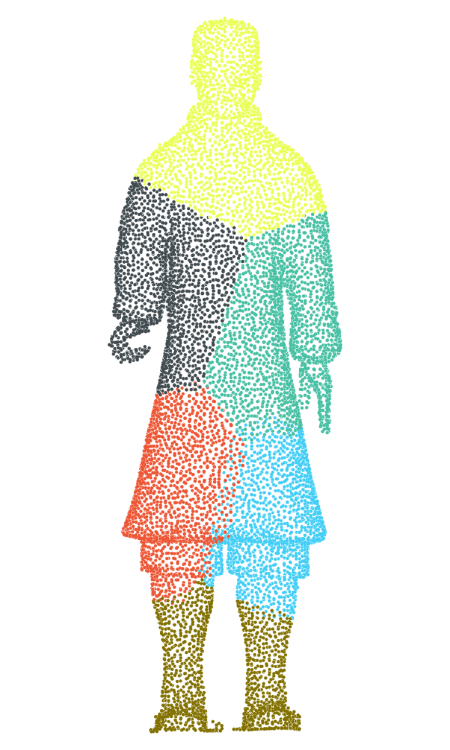} &
		\includegraphics[height=\x,keepaspectratio]{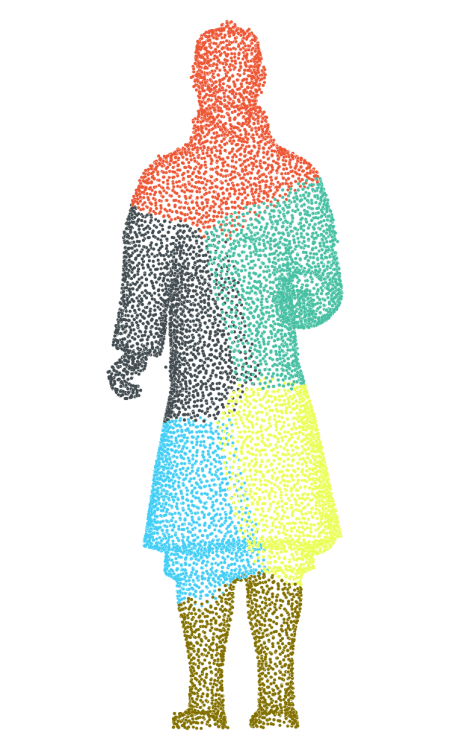} &
		\includegraphics[height=\x,keepaspectratio]{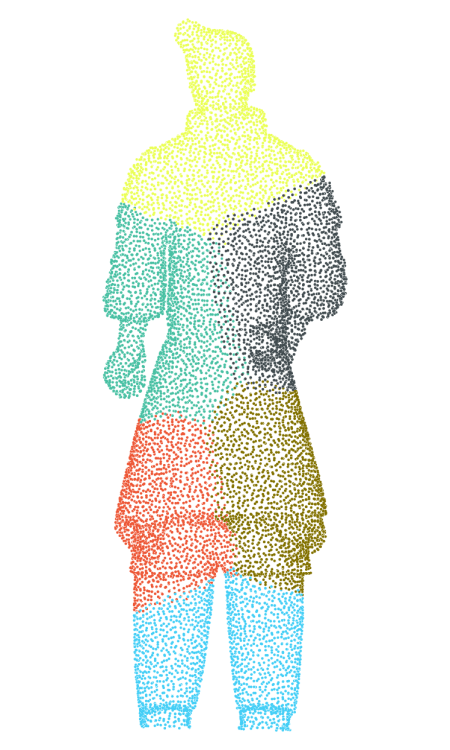} &
		\includegraphics[height=\x,keepaspectratio]{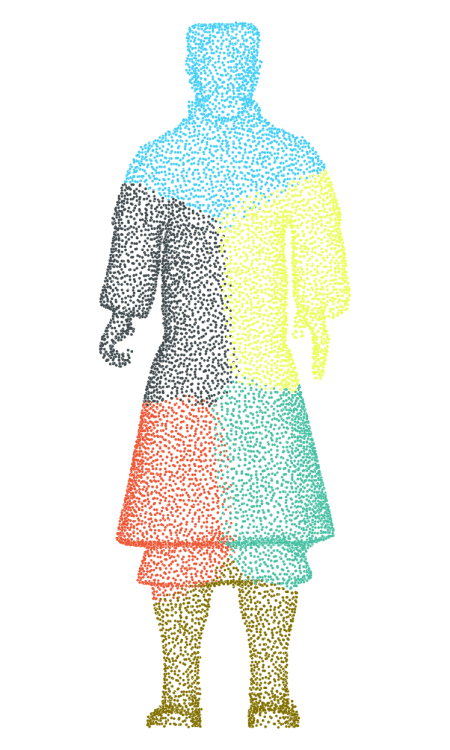}   \\
		\midrule
		SRG-Net & 
		\includegraphics[height=\x,keepaspectratio]{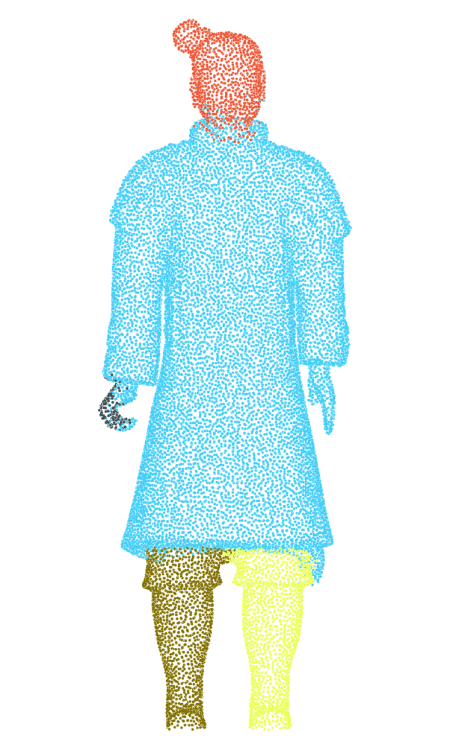} &
		\includegraphics[height=\x,keepaspectratio]{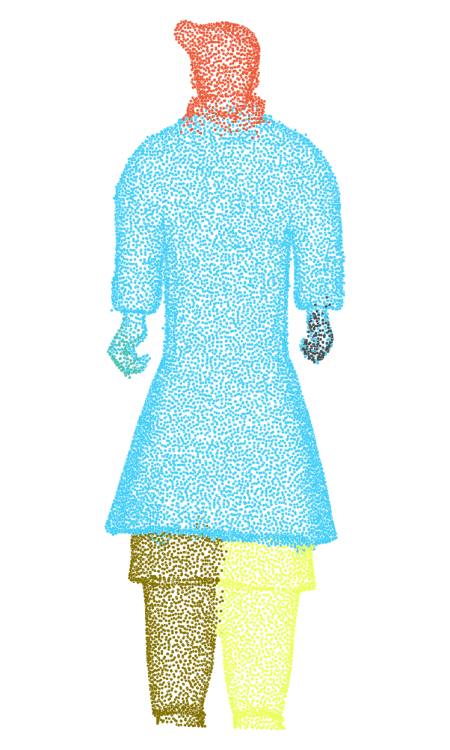} &
		\includegraphics[height=\x,keepaspectratio]{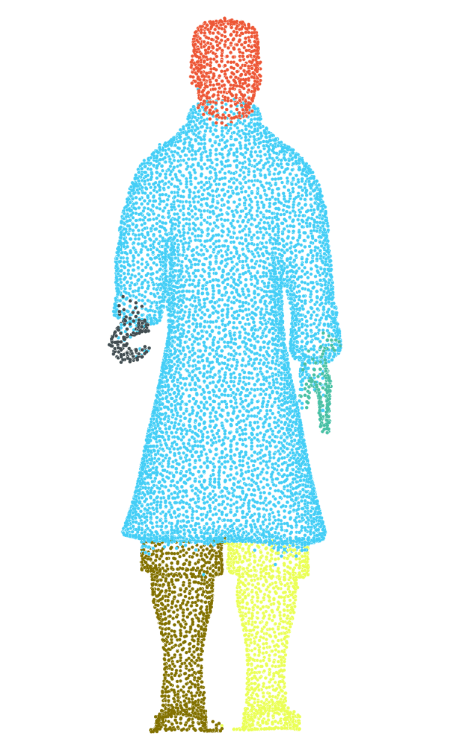} &
		\includegraphics[height=\x,keepaspectratio]{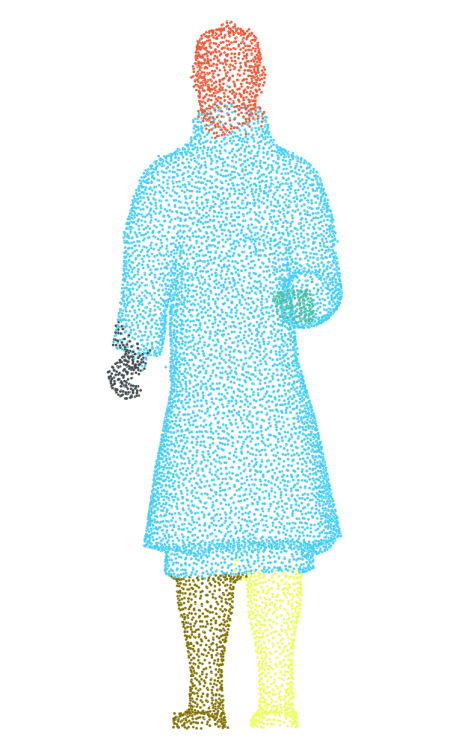} &
		\includegraphics[height=\x,keepaspectratio]{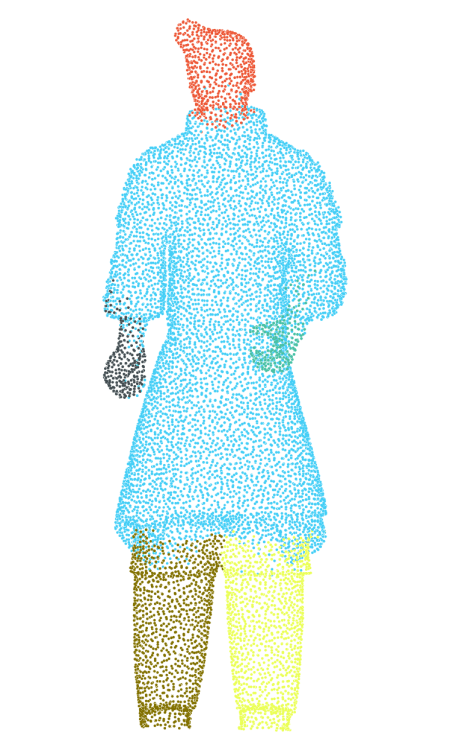} &
		\includegraphics[height=\x,keepaspectratio]{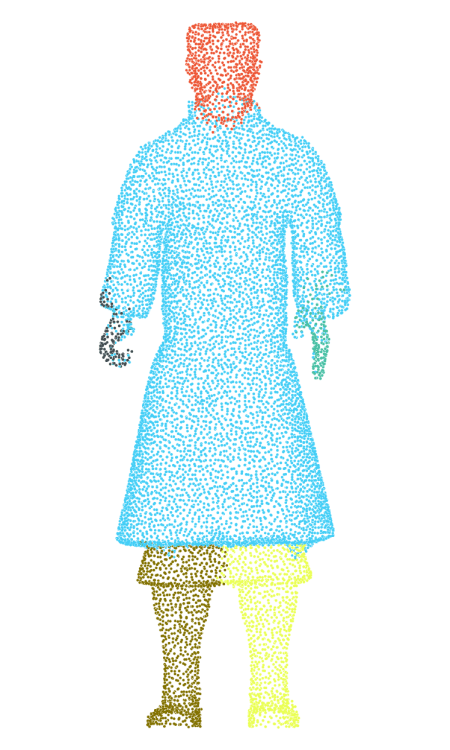}  \\
		\bottomrule
	\end{tabular}
	\caption{Visualization Results of Different Methods on Terracotta Warrior} \label{different methods figure}
\end{table*}

\subsection{Experiments on ShapeNet}

In this section, we perform experiments on ShapeNet to evaluate the robustness of our SRG-Net method. ShapeNet contains 16,881 objects from 16 categories. We set the number of the parts according to different categories separately. All models were down-sampled to 2,048 points. The evaluation metric is mean intersection-over-union (mIoU). 

Our quantitative experiment results are shown in Table~\ref{comparison shapenet}. As in Table~\ref{comparison shapenet}, SRG-Net outperforms all previous models. SRG-Net improves the overall accuracy of SRG-DGCNN by 8.2\% and even larger overhead compared with SRG-PointNet2 and SRG-PointNet. Especially, our method outperforms SRG-DGCNN on all kinds of categories, increasing 5\% accuracy on knife. Overall, our method achieves better accuracy on ShapeNet compared with other methods.

\begin{figure*}[htbp]
	\centering
	\includegraphics[width=\textwidth]{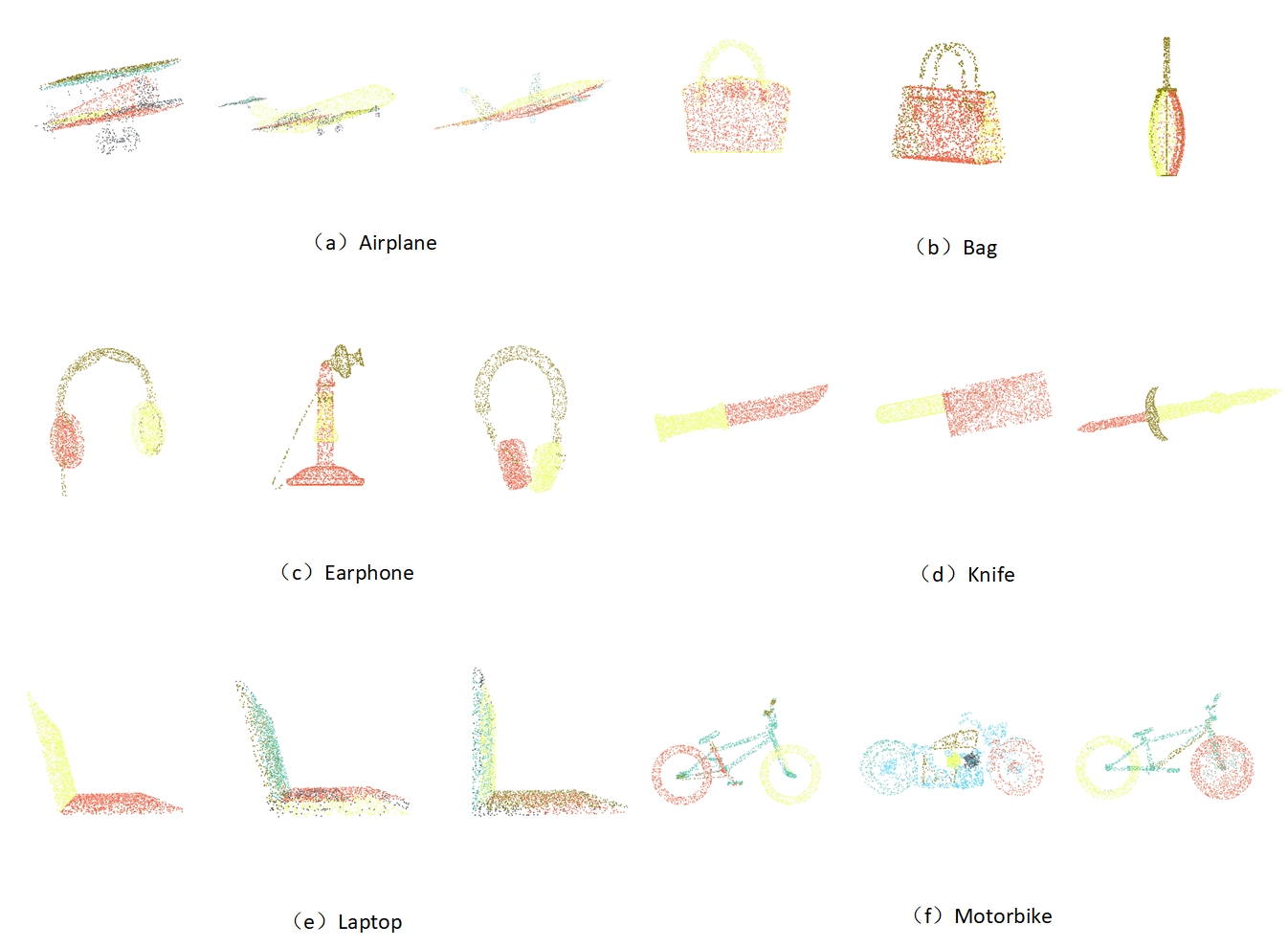}
	\caption{SRG-Net Segmentation Results of ShapeNet}
	\label{shapenet}
\end{figure*}

Some visualization results are shown in Fig.~\ref{shapenet}. As is shown in Fig.~\ref{shapenet}, SRG-Net achieves good results on bag, knife, motorbike, and achieves quite good results on airplane, earphone and laptop.

\begin{table*}[htbp]\footnotesize
    \center
	\begin{tabular}{ P{0.17\textwidth} P{0.07\textwidth} P{0.07\textwidth} P{0.07\textwidth}
	               P{0.07\textwidth} P{0.07\textwidth} P{0.07\textwidth} P{0.07\textwidth} P{0.07\textwidth}} \toprule
		{Method} & {Airplane} & {Bag} & {Earphone} & \ {Knife} & {Laptop} & {Bike} & {Mug} & {OA}\\
		\midrule
		SRG-DGCNN & 67.7393 & 61.2568 & 56.0633 & 77.0035 & 75.8854 & 70.7194 & 59.3024 & 72.6273\\ 
		SRG-Pointnet2 & 61.1865 & 54.4515 & 51.6673 & 71.9515 & 69.9074 & 64.3308 & 53.8897 & 66.4712\\ 
		SRG-Pointnet & 57.1590 & 51.0517 & 47.2101 & 67.9229 & 66.4234 & 61.2438 & 49.1265 & 61.4242\\ 
		\textbf{SRG-Net} & \textbf{73.0493} & \textbf{67.0935} & \textbf{62.4256} & \textbf{83.2430} & \textbf{82.6076} & \textbf{76.7677} & \textbf{65.8810} & \textbf{78.1954}\\ 
		\bottomrule
	\end{tabular}
	\caption{Accuracy(\%) of Different Methods on ShapeNet} \label{comparison shapenet}
\end{table*}

\subsection{Ablation Study}

In order to show the influence of different modules and epochs in our method, we conduct ablation studies on our terracotta warrior dataset, which are described in Section~\ref{SRG-Net}, which are evaluated by overall accuracy and mIoU.

\textbf{The influences of different modules.} As shown in Table~\ref{ablation study}, The results of version without the graph convolution (A, row 1) show that the network not working well in learning the topological features of the local neighborhood of the point cloud. The results of version without edge convolution (B, row 2) demonstrate that our method without edge convolution will cause the network unable to understand the relationship between points well. The results of version without refinement operation (C, row 3) reveal that the pipeline cannot set tags reasonably based on point cloud content because the number of unique cluster labels should be adaptive to context. The results show that the refinement operation increase 15.6\% on the accuracy of SRG-Net.

\begin{figure*}[htbp]
    \centering
    \subfigure[iter 250]{
        \includegraphics[width=0.10\textwidth]{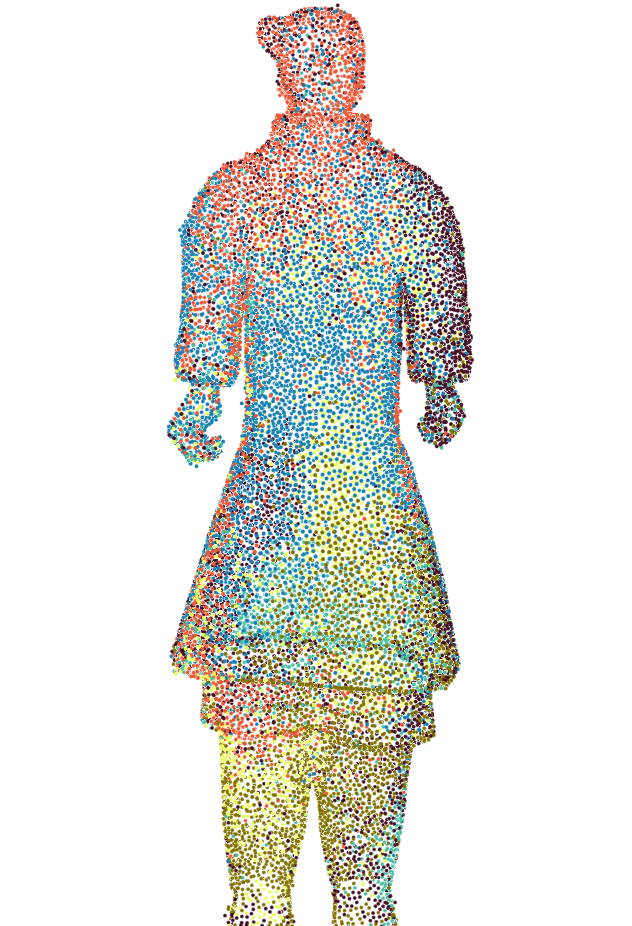}
    }
    \subfigure[iter 500]{
        \includegraphics[width=0.10\textwidth]{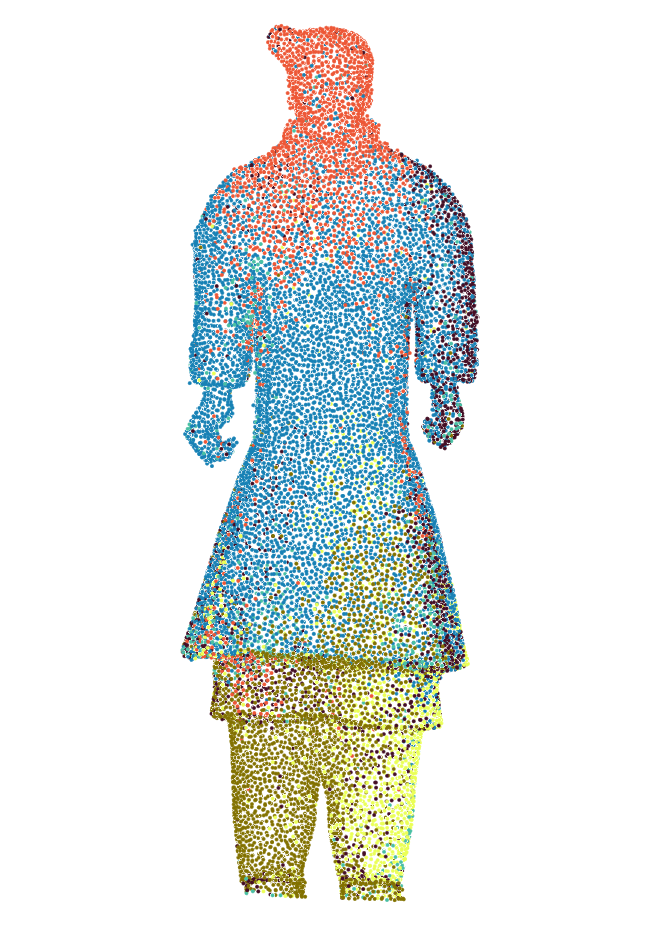}
    }
    \subfigure[iter 750]{
        \includegraphics[width=0.10\textwidth]{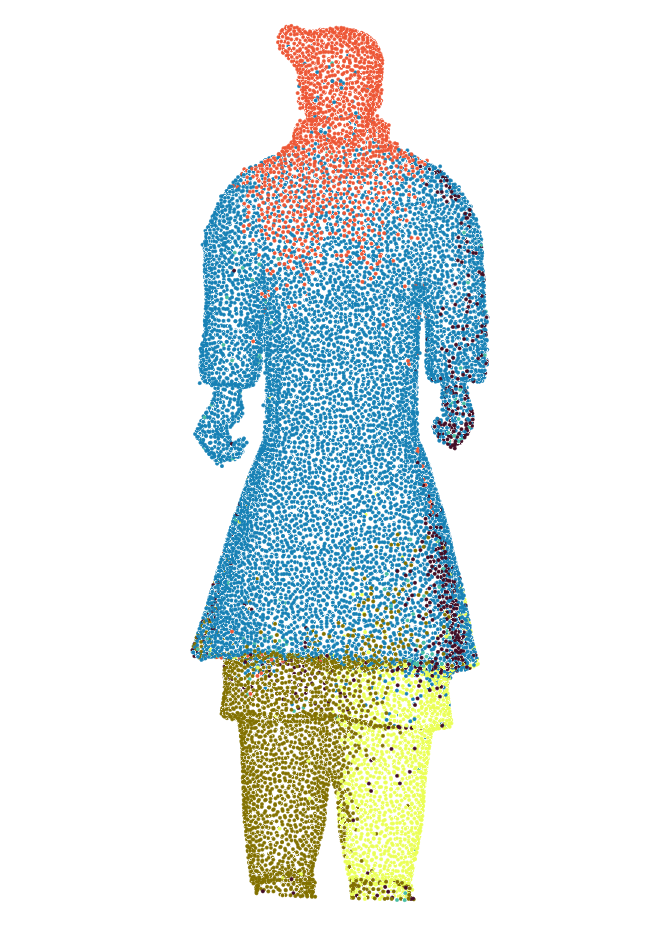}
    }
    \subfigure[iter 1000]{
        \includegraphics[width=0.10\textwidth]{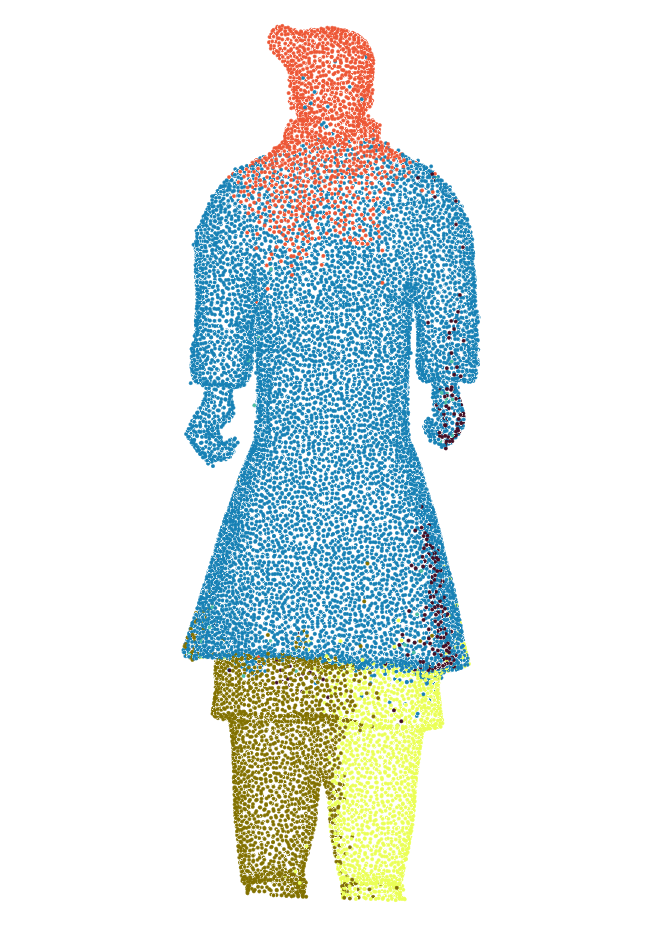}
    }
    \subfigure[iter 1250]{
        \includegraphics[width=0.10\textwidth]{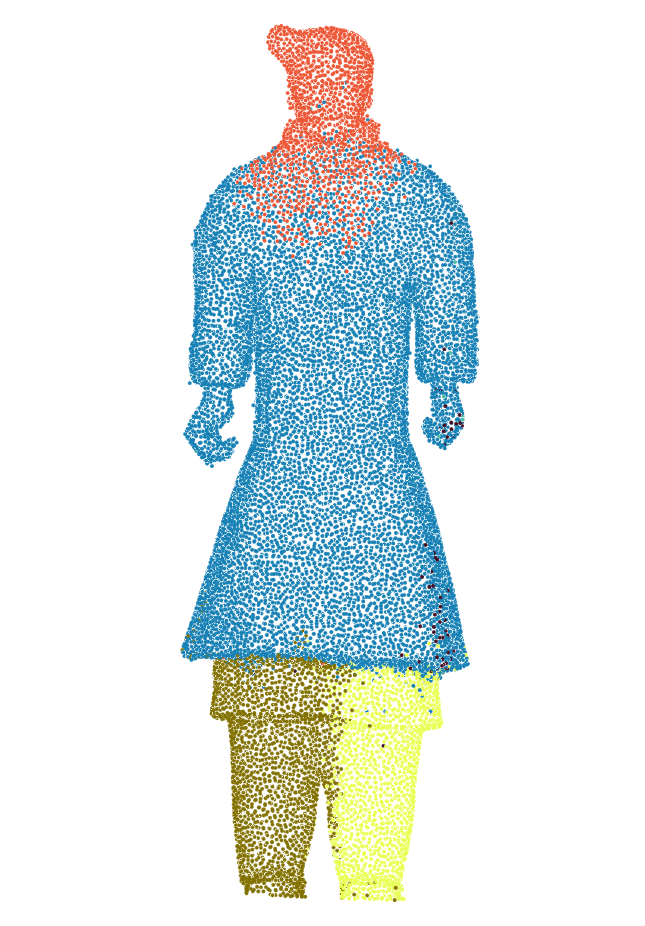}
    }
    \subfigure[iter 1500]{
        \includegraphics[width=0.10\textwidth]{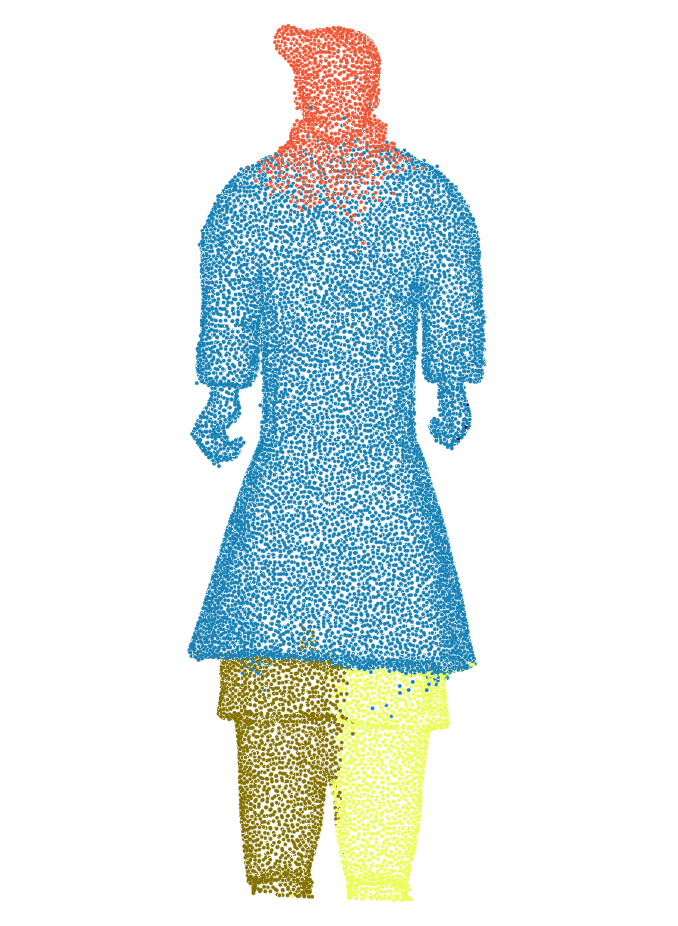}
    }
    \subfigure[iter 1750]{
        \includegraphics[width=0.10\textwidth]{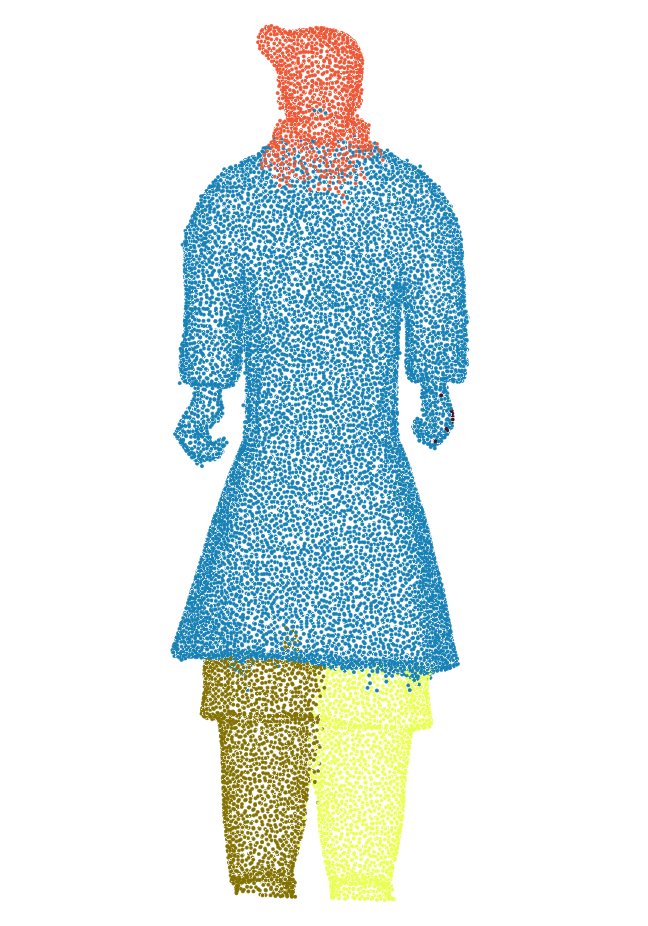}
    }
    \subfigure[iter 2000]{
        \includegraphics[width=0.10\textwidth]{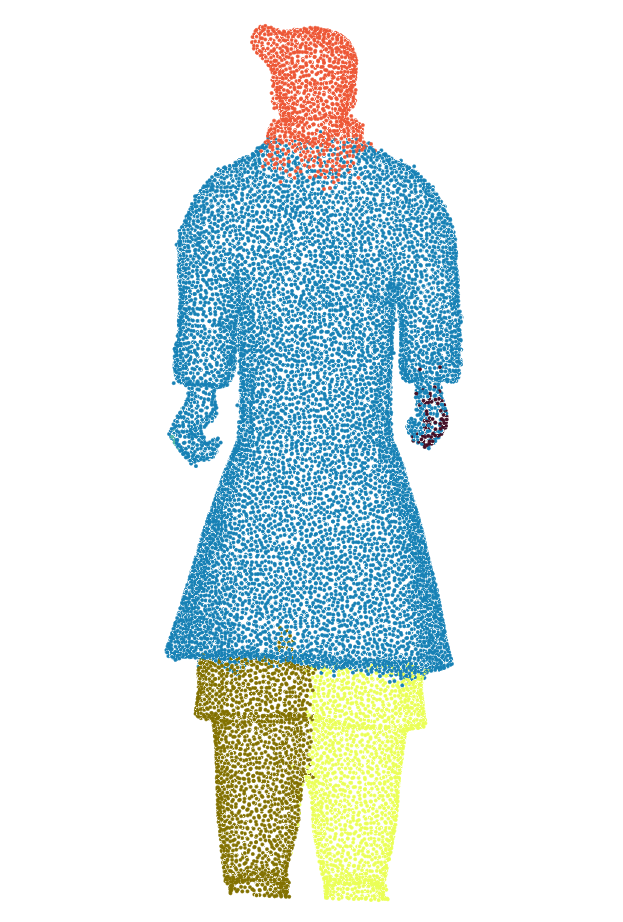}
    }
    \caption{Results of Different Iterations in SRG-Net}
    \label{figure iter}
\end{figure*}

\textbf{The influences of different epochs.} To visualize the influence of different epochs, we set the number of epochs to 2500 and set the segmentation number to 6. The segmentation results of different iterations in one terracotta warrior model are shown in Fig.~\ref{figure iter}. 

\begin{table}[htbp]\footnotesize
    \center
	\begin{tabular}{ P{0.16\textwidth} P{0.12\textwidth} P{0.12\textwidth}} \toprule
		Method & mIoU(\%) & OA(\%) \\
		\midrule
		A & 77.71 & 80.17 \\
		B & 74.18 & 76.51 \\
		C & 71.46 & 73.48 \\
		\bottomrule
	\end{tabular}
	\caption{Ablation Study} \label{ablation study}
\end{table}

\section{Conclusion}
This paper provides an end-to-end model called SRG-Net for unsupervised segmentation on the terracotta warrior point cloud. Our method aims at assisting the archaeologists in speeding up the repairing process of terracotta warriors.
Firstly, we design a novel seed-region-growing method to pre-segment point clouds with coordinates and normal value of the 3D terracotta warrior point clouds. Based on the pre-segmentation process, we proposed a CNN inspired by the dynamic graph in DG-CNN and auto-encoder in FoldingNet to learn the local and global characteristics of the 3D point cloud better. At last, we propose a refinement process and append it after the CNN process.
Finally, we evaluate our method on the terracotta warrior data, and we outperform the state-of-the-art methods with higher accuracy and less latency. Besides, we also conduct experiments on the ShapeNet, which demonstrates our approach is also useful for the human body and standard object part segmentation. 
Our work still has some limitations. The number of points from the hand of the terracotta warrior is relatively small, so it is difficult for SRG-Net to learn the characteristics of hands. We will try to work on this problem in the future. We hope our work can be helpful to the research of terracotta warriors in archaeology and point cloud work of other researchers.

\section{Acknowledgments}
This work is equally and mainly supported by the National Key Research and Development Program of China ({No. 2019YFC1521102, No. 2019YFC1521103)} and the Key Research and Development Program of Shaanxi Province ({No. 2020KW-068)}. 
Besides, this work is also partly supported by the Key Research and Development Program of Shaanxi Province (No.2019GY-215, No.2019ZDLSF07-02, No.2019ZDLGY10-01),  the Major research and development project of Qinghai(No. 2020-SF-143), the National Natural Science Foundation of China under Grant (No.61701403,No.61731015) and the Young Talent Support Program of the Shaanxi Association for Science and Technology under Grant (No.20190107).

\bibliographystyle{unsrtnat}
\bibliography{template}

\begin{thebibliography}{40}
\providecommand{\natexlab}[1]{#1}
\providecommand{\url}[1]{\texttt{#1}}
\expandafter\ifx\csname urlstyle\endcsname\relax
  \providecommand{\doi}[1]{doi: #1}\else
  \providecommand{\doi}{doi: \begingroup \urlstyle{rm}\Url}\fi

\bibitem[Tao(2020)]{tao2020}
An~Tao.
\newblock Unsupervised point cloud reconstruction for classific feature
  learning.
\newblock
  \emph{https://github.com/AnTao97/UnsupervisedPointCloudReconstruction}, 2020.

\bibitem[Yang et~al.(2018)Yang, Feng, Shen, and Tian]{yang2017foldingnet}
Yaoqing Yang, Chen Feng, Yiru Shen, and Dong Tian.
\newblock Foldingnet: Point cloud auto-encoder via deep grid deformation.
\newblock In \emph{Proceedings of the IEEE Conference on Computer Vision and
  Pattern Recognition}, pages 206--215, 2018.

\bibitem[Wang et~al.(2019)Wang, Sun, Liu, Sarma, Bronstein, and
  Solomon]{Wang_2019}
Yue Wang, Yongbin Sun, Ziwei Liu, Sanjay~E Sarma, Michael~M Bronstein, and
  Justin~M Solomon.
\newblock Dynamic graph cnn for learning on point clouds.
\newblock \emph{Acm Transactions On Graphics (tog)}, 38\penalty0 (5):\penalty0
  1--12, 2019.

\bibitem[Kruskal(1956)]{Kruskal}
Joseph~B Kruskal.
\newblock On the shortest spanning subtree of a graph and the traveling
  salesman problem.
\newblock \emph{Proceedings of the American Mathematical society}, 7\penalty0
  (1):\penalty0 48--50, 1956.

\bibitem[Lee et~al.(2009{\natexlab{a}})Lee, Pham, Largman, and
  Ng]{NIPS2009_3674}
Honglak Lee, Peter Pham, Yan Largman, and Andrew Ng.
\newblock Unsupervised feature learning for audio classification using
  convolutional deep belief networks.
\newblock \emph{Advances in neural information processing systems},
  22:\penalty0 1096--1104, 2009{\natexlab{a}}.

\bibitem[Le(2013)]{le2012building}
Quoc~V Le.
\newblock Building high-level features using large scale unsupervised learning.
\newblock In \emph{2013 IEEE international conference on acoustics, speech and
  signal processing}, pages 8595--8598. IEEE, 2013.

\bibitem[Lee et~al.(2009{\natexlab{b}})Lee, Grosse, Ranganath, and
  Ng]{10.1145/1553374.1553453}
Honglak Lee, Roger Grosse, Rajesh Ranganath, and Andrew~Y Ng.
\newblock Convolutional deep belief networks for scalable unsupervised learning
  of hierarchical representations.
\newblock In \emph{Proceedings of the 26th annual international conference on
  machine learning}, pages 609--616, 2009{\natexlab{b}}.

\bibitem[Kanezaki(2018{\natexlab{a}})]{8462533}
Asako Kanezaki.
\newblock Unsupervised image segmentation by backpropagation.
\newblock In \emph{2018 IEEE international conference on acoustics, speech and
  signal processing (ICASSP)}, pages 1543--1547. IEEE, 2018{\natexlab{a}}.

\bibitem[Achanta et~al.(2012)Achanta, Shaji, Smith, Lucchi, Fua, and
  S{\"u}sstrunk]{Achanta:177415}
Radhakrishna Achanta, Appu Shaji, Kevin Smith, Aurelien Lucchi, Pascal Fua, and
  Sabine S{\"u}sstrunk.
\newblock Slic superpixels compared to state-of-the-art superpixel methods.
\newblock \emph{IEEE transactions on pattern analysis and machine
  intelligence}, 34\penalty0 (11):\penalty0 2274--2282, 2012.

\bibitem[Kim et~al.(2020)Kim, Kanezaki, and Tanaka]{Kim_2020}
Wonjik Kim, Asako Kanezaki, and Masayuki Tanaka.
\newblock Unsupervised learning of image segmentation based on differentiable
  feature clustering.
\newblock \emph{IEEE Transactions on Image Processing}, 29:\penalty0
  8055--8068, 2020.

\bibitem[Lawin et~al.(2017)Lawin, Danelljan, Tosteberg, Bhat, Khan, and
  Felsberg]{lawin2017deep}
Felix~J{\"a}remo Lawin, Martin Danelljan, Patrik Tosteberg, Goutam Bhat,
  Fahad~Shahbaz Khan, and Michael Felsberg.
\newblock Deep projective 3d semantic segmentation.
\newblock In \emph{International Conference on Computer Analysis of Images and
  Patterns}, pages 95--107. Springer, 2017.

\bibitem[Boulch et~al.(2017)Boulch, Le~Saux, and
  Audebert]{10.2312/3dor.20171047}
Alexandre Boulch, Bertrand Le~Saux, and Nicolas Audebert.
\newblock Unstructured point cloud semantic labeling using deep segmentation
  networks.
\newblock \emph{3DOR}, 2:\penalty0 7, 2017.

\bibitem[Wu et~al.(2018)Wu, Wan, Yue, and
  Keutzer]{DBLP:journals/corr/abs-1710-07368}
Bichen Wu, Alvin Wan, Xiangyu Yue, and Kurt Keutzer.
\newblock Squeezeseg: Convolutional neural nets with recurrent crf for
  real-time road-object segmentation from 3d lidar point cloud.
\newblock In \emph{2018 IEEE International Conference on Robotics and
  Automation (ICRA)}, pages 1887--1893. IEEE, 2018.

\bibitem[Milioto et~al.(2019)Milioto, Vizzo, Behley, and Stachniss]{8967762}
Andres Milioto, Ignacio Vizzo, Jens Behley, and Cyrill Stachniss.
\newblock Rangenet++: Fast and accurate lidar semantic segmentation.
\newblock In \emph{2019 IEEE/RSJ International Conference on Intelligent Robots
  and Systems (IROS)}, pages 4213--4220. IEEE, 2019.

\bibitem[Graham et~al.(2018)Graham, Engelcke, and Van
  Der~Maaten]{DBLP:journals/corr/abs-1711-10275}
Benjamin Graham, Martin Engelcke, and Laurens Van Der~Maaten.
\newblock 3d semantic segmentation with submanifold sparse convolutional
  networks.
\newblock In \emph{Proceedings of the IEEE conference on computer vision and
  pattern recognition}, pages 9224--9232, 2018.

\bibitem[Dai and Nie{\ss}ner(2018)]{dai20183dmv}
Angela Dai and Matthias Nie{\ss}ner.
\newblock 3dmv: Joint 3d-multi-view prediction for 3d semantic scene
  segmentation.
\newblock In \emph{Proceedings of the European Conference on Computer Vision
  (ECCV)}, pages 452--468, 2018.

\bibitem[Jaritz et~al.(2019)Jaritz, Gu, and Su]{jaritz2019multiview}
Maximilian Jaritz, Jiayuan Gu, and Hao Su.
\newblock Multi-view pointnet for 3d scene understanding.
\newblock In \emph{Proceedings of the IEEE International Conference on Computer
  Vision Workshops}, pages 0--0, 2019.

\bibitem[Wang et~al.(2018{\natexlab{a}})Wang, Suo, Ma, Pokrovsky, and
  Urtasun]{8578372_deep_parametric}
Shenlong Wang, Simon Suo, Wei-Chiu Ma, Andrei Pokrovsky, and Raquel Urtasun.
\newblock Deep parametric continuous convolutional neural networks.
\newblock In \emph{Proceedings of the IEEE Conference on Computer Vision and
  Pattern Recognition}, pages 2589--2597, 2018{\natexlab{a}}.

\bibitem[Qi et~al.(2017{\natexlab{a}})Qi, Su, Mo, and Guibas]{Charles_2017}
Charles~R Qi, Hao Su, Kaichun Mo, and Leonidas~J Guibas.
\newblock Pointnet: Deep learning on point sets for 3d classification and
  segmentation.
\newblock In \emph{Proceedings of the IEEE conference on computer vision and
  pattern recognition}, pages 652--660, 2017{\natexlab{a}}.

\bibitem[Qi et~al.(2017{\natexlab{b}})Qi, Yi, Su, and Guibas]{qi2017pointnet}
Charles~Ruizhongtai Qi, Li~Yi, Hao Su, and Leonidas~J Guibas.
\newblock Pointnet++: Deep hierarchical feature learning on point sets in a
  metric space.
\newblock In \emph{Advances in neural information processing systems}, pages
  5099--5108, 2017{\natexlab{b}}.

\bibitem[Jiang et~al.(2018)Jiang, Wu, Zhao, Zhao, and
  Lu]{DBLP:journals/corr/abs-1807-00652}
Mingyang Jiang, Yiran Wu, Tianqi Zhao, Zelin Zhao, and Cewu Lu.
\newblock Pointsift: A sift-like network module for 3d point cloud semantic
  segmentation.
\newblock \emph{arXiv preprint arXiv:1807.00652}, 2018.

\bibitem[Zhao et~al.(2019)Zhao, Jiang, Fu, and Jia]{Zhao_2019_CVPR}
Hengshuang Zhao, Li~Jiang, Chi-Wing Fu, and Jiaya Jia.
\newblock Pointweb: Enhancing local neighborhood features for point cloud
  processing.
\newblock In \emph{Proceedings of the IEEE Conference on Computer Vision and
  Pattern Recognition}, pages 5565--5573, 2019.

\bibitem[Wu et~al.(2019)Wu, Qi, and Fuxin]{wu2019pointconv}
Wenxuan Wu, Zhongang Qi, and Li~Fuxin.
\newblock Pointconv: Deep convolutional networks on 3d point clouds.
\newblock In \emph{Proceedings of the IEEE Conference on Computer Vision and
  Pattern Recognition}, pages 9621--9630, 2019.

\bibitem[Li et~al.(2018)Li, Bu, Sun, Wu, Di, and Chen]{li2018pointcnn}
Yangyan Li, Rui Bu, Mingchao Sun, Wei Wu, Xinhan Di, and Baoquan Chen.
\newblock Pointcnn: Convolution on $\mathcal{X}$-transformed points, 2018.

\bibitem[Wang et~al.(2018{\natexlab{b}})Wang, Suo, Ma, Pokrovsky, and
  Urtasun]{8578372}
Shenlong Wang, Simon Suo, Wei-Chiu Ma, Andrei Pokrovsky, and Raquel Urtasun.
\newblock Deep parametric continuous convolutional neural networks.
\newblock In \emph{Proceedings of the IEEE Conference on Computer Vision and
  Pattern Recognition}, pages 2589--2597, 2018{\natexlab{b}}.

\bibitem[Veli{\v{c}}kovi{\'c} et~al.(2017)Veli{\v{c}}kovi{\'c}, Cucurull,
  Casanova, Romero, Lio, and Bengio]{2018graph}
Petar Veli{\v{c}}kovi{\'c}, Guillem Cucurull, Arantxa Casanova, Adriana Romero,
  Pietro Lio, and Yoshua Bengio.
\newblock Graph attention networks.
\newblock \emph{arXiv preprint arXiv:1710.10903}, 2017.

\bibitem[Wang and Lu(2019)]{wang2019voxsegnet}
Zongji Wang and Feng Lu.
\newblock Voxsegnet: Volumetric cnns for semantic part segmentation of 3d
  shapes.
\newblock \emph{IEEE transactions on visualization and computer graphics},
  2019.

\bibitem[Yi et~al.(2017)Yi, Su, Guo, and Guibas]{yi2016syncspeccnn}
Li~Yi, Hao Su, Xingwen Guo, and Leonidas~J Guibas.
\newblock Syncspeccnn: Synchronized spectral cnn for 3d shape segmentation.
\newblock In \emph{Proceedings of the IEEE Conference on Computer Vision and
  Pattern Recognition}, pages 2282--2290, 2017.

\bibitem[Liu and Xiong(2008)]{LIU2008576}
Yu~Liu and Youlun Xiong.
\newblock Automatic segmentation of unorganized noisy point clouds based on the
  gaussian map.
\newblock \emph{Computer-Aided Design}, 40\penalty0 (5):\penalty0 576--594,
  2008.

\bibitem[Di~Angelo and Di~Stefano(2015)]{DIANGELO201544}
Luca Di~Angelo and Paolo Di~Stefano.
\newblock Geometric segmentation of 3d scanned surfaces.
\newblock \emph{Computer-Aided Design}, 62:\penalty0 44--56, 2015.

\bibitem[Benk{\H{o}} and V{\'a}rady(2004)]{BENKO2004511}
P{\'a}l Benk{\H{o}} and Tam{\'a}s V{\'a}rady.
\newblock Segmentation methods for smooth point regions of conventional
  engineering objects.
\newblock \emph{Computer-Aided Design}, 36\penalty0 (6):\penalty0 511--523,
  2004.

\bibitem[Chen et~al.(2019)Chen, Yin, Fisher, Chaudhuri, and
  Zhang]{chen2019baenet}
Zhiqin Chen, Kangxue Yin, Matthew Fisher, Siddhartha Chaudhuri, and Hao Zhang.
\newblock Bae-net: Branched autoencoder for shape co-segmentation.
\newblock In \emph{Proceedings of the IEEE International Conference on Computer
  Vision}, pages 8490--8499, 2019.

\bibitem[OuYang and Feng(2005)]{OUYANG20051071}
Daoshan OuYang and Hsi-Yung Feng.
\newblock On the normal vector estimation for point cloud data from smooth
  surfaces.
\newblock \emph{Computer-Aided Design}, 37\penalty0 (10):\penalty0 1071--1079,
  2005.

\bibitem[Zhou et~al.(2020)Zhou, Huang, Liu, and Liu]{ZHOU2020102916}
Jun Zhou, Hua Huang, Bin Liu, and Xiuping Liu.
\newblock Normal estimation for 3d point clouds via local plane constraint and
  multi-scale selection.
\newblock \emph{Computer-Aided Design}, 129:\penalty0 102916, 2020.

\bibitem[Wang et~al.(2013)Wang, Feng, Delorme, and Engin]{WANG20131333}
Yutao Wang, Hsi-Yung Feng, F{\'e}lix-{\'E}tienne Delorme, and Serafettin Engin.
\newblock An adaptive normal estimation method for scanned point clouds with
  sharp features.
\newblock \emph{Computer-Aided Design}, 45\penalty0 (11):\penalty0 1333--1348,
  2013.

\bibitem[Rusu and Cousins(2011)]{Normal_Estimation}
Radu~Bogdan Rusu and Steve Cousins.
\newblock 3d is here: Point cloud library (pcl).
\newblock In \emph{2011 IEEE international conference on robotics and
  automation}, pages 1--4. IEEE, 2011.

\bibitem[Kanezaki(2018{\natexlab{b}})]{Kanezaki}
Asako Kanezaki.
\newblock Unsupervised image segmentation by backpropagation.
\newblock In \emph{2018 IEEE international conference on acoustics, speech and
  signal processing (ICASSP)}, pages 1543--1547. IEEE, 2018{\natexlab{b}}.

\bibitem[Europe(2020)]{artec}
Artec Europe.
\newblock 3d object scanner artec eva | best structured-light 3d scanning
  device, 2020.
\newblock URL \url{https://www.artec3d.com/portable-3d-scanners/artec-eva}.

\bibitem[Zhang et~al.(2020)Zhang, Wang, Kadam, Liu, and
  Kuo]{zhang2020pointhop++}
Min Zhang, Yifan Wang, Pranav Kadam, Shan Liu, and C-C~Jay Kuo.
\newblock Pointhop++: A lightweight learning model on point sets for 3d
  classification.
\newblock \emph{arXiv preprint arXiv:2002.03281}, 2020.

\bibitem[Simonovsky and Komodakis(2017)]{Simonovsky2017ecc}
Martin Simonovsky and Nikos Komodakis.
\newblock Dynamic edge-conditioned filters in convolutional neural networks on
  graphs.
\newblock \emph{https://arxiv.org/abs/1704.02901}, 2017.

\end{thebibliography}

\end{document}